\pdfoutput=1

\documentclass[11pt]{article}

\usepackage{ACL2023}

\usepackage{times}
\usepackage{latexsym}
\usepackage{amsmath}

\usepackage[T1]{fontenc}

\usepackage[utf8]{inputenc}

\usepackage{microtype}

\usepackage{inconsolata}

\usepackage{graphicx} 

\title{Vision Language Transformers: A Survey}

\author{Clayton Fields \\
  Boise State University \\ 
  1910 W University Dr \\
  Boise, ID 83725 \\
  \texttt{claytonfields@u.boisestate.edu} \\\And
  Casey Kennington \\
  Boise State University \\ 
  1910 W University Dr \\
  Boise, ID 83725 \\
  \texttt{caseykennington@boisestate.edu} \\}

\begin{document}
\maketitle
\begin{abstract}
Vision language tasks, such as answering questions about or generating captions that describe an image, are difficult tasks for computers to perform. A relatively recent body of research has adapted the pretrained transformer architecture introduced in \citet{vaswani2017attention} to vision language modeling. Transformer models have greatly improved performance and versatility over previous vision language models. They do so by pretraining models on a large generic datasets and transferring their learning to new tasks with minor changes in architecture and parameter values. This type of transfer learning has become the standard modeling practice in both natural language processing and computer vision. Vision language transformers offer the promise of producing similar advancements in tasks which require both vision and language. In this paper, we provide a broad synthesis of the currently available research on vision language transformer models and offer some analysis of their strengths, limitations and some open questions that remain.
\end{abstract}

\section{Introduction}
\label{sec:intro} 

Vision language modeling is the domain where computer vision and natural language processing intersect. An example of a VL task is visual question answering: given an image and a question about an image, a VL model must choose the correct answer out of a number of choices. Another example, and a more challenging task, is image captioning, given an image a model must produce a sequence of text describing the picture. Though effortless for human beings, tasks of this nature have historically proven extremely challenging for computers to perform. Until fairly recently, deep learning models for VL tasks tended to be both conceptually convoluted and confined to a narrow range of uses. 

In the past few years a new class of models called VL transformers have greatly expanded the accuracy and versatility of vision language models. These models are based on the celebrated transformer architecture introduced in \citet{vaswani2017attention}. VL transformers improve on the previous paradigms by pretraining models on large datasets of image-text pairs before transferring them to other tasks, usually with minor changes to parameter values and architecture. In a very short space of time a bewildering array of these models have appeared in the literature. They vary widely in their intended uses, architectures, pretraining processes as well as the data used to pretrain them. 

In this paper, we provide a comprehensive survey of the various VL transformer models found in the literature. These models were designed for a wide range of vision language tasks. Models such as CLIP \cite{radford2021learning} and ALIGN \cite{jia2021scaling} are particularly well suited to vision language alignment tasks such image retrieval. Whereas models like UNITER \cite{Chen2019-sh}, ViLBERT \cite{Lu2019-bb} and METER \cite{dou2022empirical} specialize in understanding tasks such as visual question answering (VQA) described in the introductory paragraph. Transformers with suitable architectures, LEMON \cite{Hu2022-kq} and GIT \cite{wang2022git} to name two, were designed to generate text such as captions for image inputs. There is even a series of VL transformers specializing in visual grounding tasks in which a model must match words to the visual objects they describe. Referring Transformer and mDETR are two such models that can perform object detection on image inputs and match these objects to text descriptions. 

In the interest of brevity, we will restrict our study to models using English as their primary language. This excludes not only models for text in other languages, but also multi-lingual models. We also exclude models that are designed exclusively for video-language tasks. It should be noted however that some of the models we reviewed process video inputs as well as images. And one multi-lingual model, PaLI \cite{chen2022pali}, is included because of its superior performance on english language VL benchmarks. 

The impressive range of tasks mentioned above is reflected by an equally impressive variety of embedding strategies, model architectures, pretraining tasks and training datasets. We will discuss these topics in some detail as well as the various ways these features can be adapted to the VL domain. Along the way we hope to provide some insight into the various design choices of these models and when sufficient data exists, the corresponding effects on their performance. All of the models reviewed for this paper are listed in Table~\ref{table:models} along with the references for the papers introducing each model and some basic information about there design.  

The remainder of the paper is organized as follows, in Section~\ref{sec:transformers}, we provide a brief explanation of the transformer model that forms the basis of the models that we have reviewed and how pretrained trasformers have been adapted for NLP and CV tasks. In Section~\ref{sec:embedding} we discuss how VL models embed visual and linguistic data into their feature space, with special attention paid to how they create visual features. Section~\ref{sec:architecture} addresses the architectures of the reviewed models and how these design choices affect the interactions of visual and linguistic features. The various pretraining tasks and strategies these models use and how they affect downstream performance are summarized in Section~\ref{sec:pretraining}. Section~\ref{sec:capability} describes the models downstream capabilities and Section~\ref{sec:data} describes the data used for pretraining. In the final section we provide a brief analysis of the strengths and limitations of the models discussed and explore some future directions for research and identify open questions that remain.

\begin{table*}
\centering
\begin{tabular}{|l|l|l|l|}
 \hline
 \textbf{Model} & \textbf{Source Paper} & \textbf{Architecture} &\textbf{Visual Embedding}  \\ 
 \hline
 \hline
 \textbf{ALBEF} & \citet{li2021align} & Combo. Encoder & Patch Embeddings\\ 
 \hline
 \textbf{ALIGN} & \citet{jia2021scaling}  & Dual Encoder & Grid Features\\ 
 \hline
 \textbf{BEiT-3} & \citet{wang2023image} & Combo. Encoder & Patch Embeddings\\ 
 \hline
 \textbf{BLIP-2} & \citet{li2023blip}  & Encoder-Decoder & Patch Embeddings\\ 
 \hline
 \textbf{BridgeTower} & \citet{Xu2022-ce} & Two-Tower Encoder & Patch Embeddings\\ 
 \hline
 \textbf{CLIP} & \citet{radford2021learning} & Dual Encoder & Grid Features \\ 
 \hline
 \textbf{CoCa} & \citet{Yu2022-zn} & Encoder-Decoder & Patch Embeddings \\ 
 \hline
 \textbf{DaVinci} & \citet{diao2022prefix} & Encoder-Decoder &	Patch Embeddings \\ 
 \hline
 \textbf{DQ-DETR} & \citet{Liu2022-pg} & Encoder-Decoder & Grid Features \\ 
 \hline
 \textbf{E2E-VLP} &  \citet{xu2021e2e} & Encoder-Decoder & Grid Features\\
 \hline
 \textbf{Flamingo} & \citet{Alayrac2022-yp} & Encoder-Decoder & Grid Features \\ 
 \hline
 \textbf{FLAVA} & \citet{singh2022flava} & Combo. Encoder & Patch Embeddings\\
 \hline
 \textbf{Florence} & \citet{Yuan2021-yg} & Dual Encoder & Grid Features\\ 
 \hline
 \textbf{GIT} & \citet{wang2022git} & Encoder-Decoder & Grid Features \\
 \hline
 \textbf{GPV} & \citet{Gupta2021-kn} & Encoder-Decoder & Grid Features\\
 \hline
 \textbf{KD-VLP} & \citet{Liu2022-qj} & One-Tower Encoder & Grid Features \\ 
 \hline
 \textbf{LEMON} & \citet{Hu2022-kq} 	& Encoder-Decoder & Region Features \\ 
 \hline
 \textbf{Lit} & \citet{zhai2022lit} 	& Dual Encoder & Grid Features \\ 
 \hline
 \textbf{LXMERT} & \citet{Tan2019-yi} & Two-Tower Encoder & Region Features \\ 
 \hline
 \textbf{mDETR} & \citet{kamath2021mdetr} & Encoder-Decoder & Grid Features\\ 
 \hline
 \textbf{METER} & \citet{dou2022empirical} & Two-Tower Encoder & Patch Embeddings\\ 
 \hline
 \textbf{mPLUG} & \citet{li2022mplug} & Encoder-Decoder & Patch Embeddings\\ 
 \hline
 \textbf{OFA} & \citet{Wang2022-gx} & Encoder-Decoder & Patch Embeddings \\ 
 \hline
 \textbf{OmniVL} & \citet{wang2022omnivl} & Encoder-Decoder & Patch Embeddings\\ 
 \hline
 \textbf{Oscar} & \citet{li2020oscar} & One-Tower Encoder & Region Features\\ 
 \hline
 \textbf{PaLI} & \citet{chen2022pali} & Encoder-Decoder & Patch Embeddings \\ 
 \hline
 \textbf{PixelBERT} & \citet{huang2020pixel} & One-Tower Encoder & Grid Features \\ 
 \hline
 \textbf{Referring Transformer} & \citet{Li2021-wn} & Encoder-Decoder & Grid Features\\ 
 \hline
 \textbf{SimVLM} & \citet{Wang2021-se} & Encoder-Decoder & Patch Embeddings \\ 
 \hline
 \textbf{SOHO} & \citet{Huang2021-ee} & One-Tower Encoder & Grid Features	 \\ 
 \hline
 \textbf{Unicoder-VL} & \citet{Li2020-ea} & One-Tower Encoder & Region Features \\ 
 \hline
 \textbf{UNIMO} & \citet{Li2020-up} & One-Tower Encoder & Region Features	 \\ 
 \hline
 \textbf{UniTAB} & \citet{Yang2022-hf} & Encoder-Decoder & Grid Features \\ 
 \hline
 \textbf{UNITER} & \citet{Chen2019-sh} & One-Tower Encoder & Region Features \\ 
 \hline
 \textbf{ViLBERT} & \citet{Lu2019-bb} & Two-Tower Encoder & Region Features	 \\ 
  \hline
 \textbf{VILLA} & \citet{gan2020large} & One-Tower Encoder & Region Features	 \\ 
 \hline
 \textbf{ViLT} & \citet{kim2021vilt} & One-Tower Encoder & Patch Embeddings \\ 
 \hline
 \textbf{VinVL} & \citet{zhang2021vinvl} & One-Tower Encoder & Region Features \\ 
 \hline
 \textbf{VisualBERT} & \citet{li2019visualbert} & One-Tower Encoder & Region Features \\ 
 \hline
 \textbf{VL-BERT} & \citet{Su2019-pf}& One-Tower Encoder & Region Features \\ 
 \hline
 \textbf{VLMo} & \citet{bao2022vlmo} & Combo. Encoder & Patch Embeddings\\ 
 \hline
 \textbf{VL-T5} & \citet{cho2021unifying} & Encoder-Decoder & Region Features \\ 
 \hline
 \textbf{X2-VLM} & \citet{Zeng2022-bp} 	& Combo. Encoder & Patch Embeddings \\ 
 \hline
\end{tabular}
\caption{\label{small-model-test}
List of the models reviewed for this paper, the source paper that introduced the model and the architecture and visual embedding strategy it uses.
}
\label{table:models}
\end{table*}

\section{Background: Transformers}
\label{sec:transformers}

In this section, we describe the transformer-style deep neural models that form the architectural basis of the VL models we discuss below. Transformers were first introduced in the seminal paper "Attention Is All You Need" by \citet{vaswani2017attention} in the context of using attention mechanisms for machine translation tasks. Since then, transformers have replaced recurrent neural networks (RNN) as the standard model for most NLP tasks. NLP transformers have achieved their remarkable results by pretraining networks  with large unlabelled sets of text and then transferring the pretrained networks to other tasks with small changes in architecture and minimal parameter updates. Pretrained transformer models, such as RoBERTa \cite{liu2019roberta} and GPT-3 \cite{brown2020language}, are now state-of-the-art in virtually every category of NLP tasks.

Convolutional neural networks (CNN) are still widely used for CV tasks as of the writing of this paper. However, recent efforts have shown that the transformer architecture can be adapted to CV tasks with relatively few modifications, \cite{dosovitskiy2020image, touvron2021training}. When pretrained with a sufficiently large dataset, vision transformers can perform competitively with state-of-the-art CNNs whose architectures were designed for CV.

Given their ability to perform at or near state-of-the-art in both domains, transformers have became the natural choice as the basis for pretrained VL models. Before we move on to discussing the design choices involved in adapting transformers to VL tasks, we will offer a brief overview of the transformer model and the attention mechanism that powers its remarkable results. Readers well versed in the working of transformers and their applications NLP and CV should feel free to proceed to the next section.

\subsection{Architecture of Transformers}
\label{subsec:trans-architecture}

In this subsection we describe the architecture of the original transformer model. Of particular interest is the self-attention mechanism that underlies the model's success in sequence processing. Unless otherwise stated, the source for the exposition in this subsection is \citet{vaswani2017attention}.

\subsubsection{Encoder and Decoder Stacks}

The first implementation of the transformer models was based on an encoder-decoder design. Formally, a given input sequence of symbol representations $\mathbf{x} = (x_1, ..., x_n)$ is mapped to an intermediate representation $\mathbf{z} = (z_1, ..., z_n)$ by the encoder module. With $\mathbf{z}$ as input, the decoder generates an output sequence $\mathbf{y} = (y_1, ..., y_n)$.

The transformer encoder stack consists of $N$ transformer layers of identical dimension. Each transformer layer, in turn, consists of a multi-head attention (MHA) sub-layer and a feed-forward network sub-layer (FFN), both of which are described in the following sub-sections. A residual connection \cite{He2015-et} around each sub-layer is then followed by a layer normalization. Formally this amounts to LayerNorm($\mathbf{x}$ + Sublayer($\mathbf{x}$)), where Sublayer($\mathbf{x}$) is the MHA or FFN sublayer function itself.

The decoder is also a stack of $N$ layers of identical dimension. However, layers in the decoder stack contain a third multi-head attention sub-layer that attends to the output of the encoder stack. The self-attention mechanism in the decoder stack is also modified so that previous positions in the sequence are masked. Masking combined with an offset applied to the output embeddings ensure that decoder predictions are based only on previous outputs. These modifications make the decoder well suited to generative tasks. In the next subsection we describe the multi-head attention mechanism key to the transformer's operation.

\begin{figure}[t]
\centering
\includegraphics[width=9cm, height=12cm]{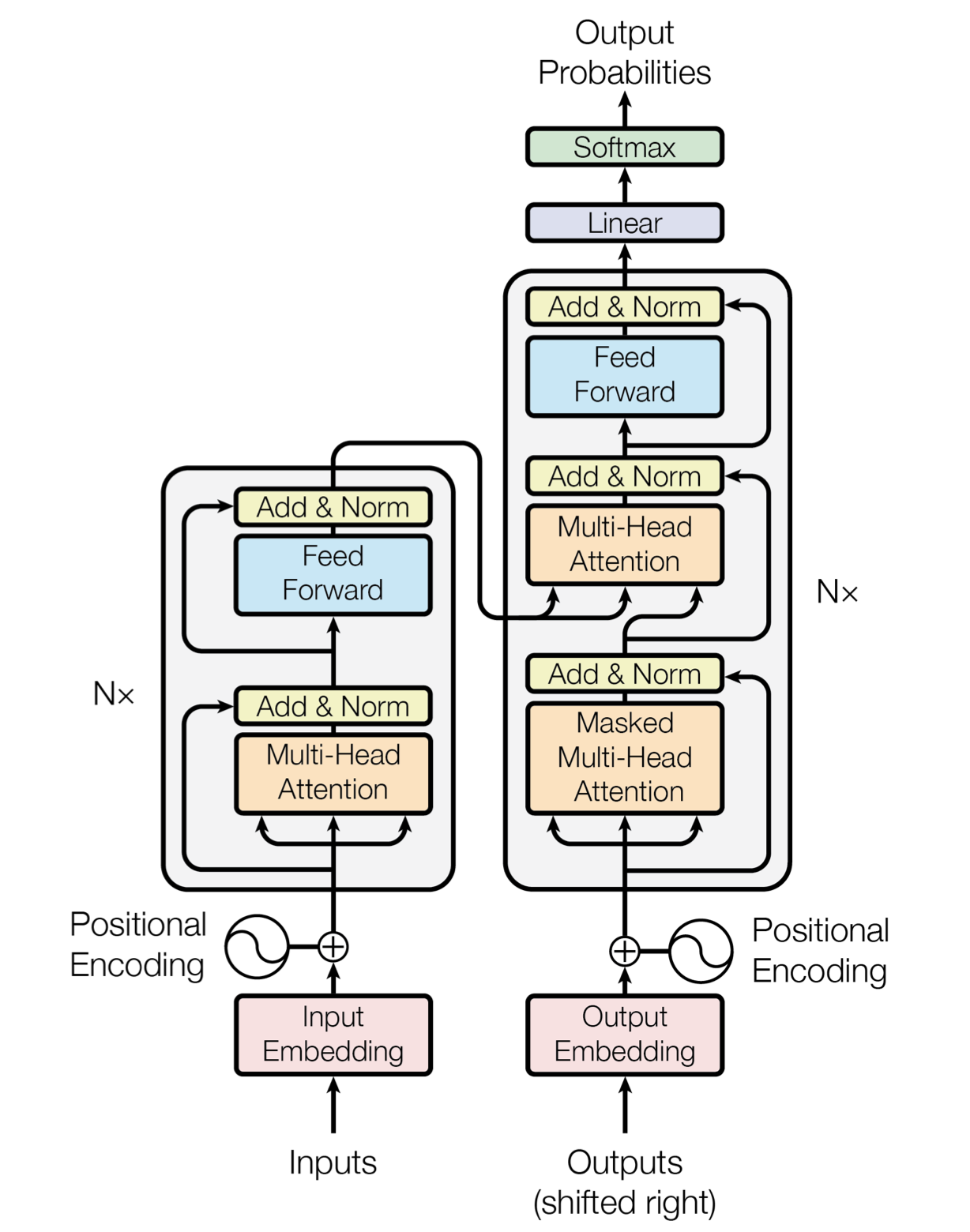}
\caption{The transformer model architecture. From \citet{vaswani2017attention}}
\label{fig:transformer-architecture}
\end{figure}

\subsubsection{Multi-Head Attention Sub-Layer}

Broadly speaking, an attention mechanism is a function that maps a query vector and a set of key-value vector pairs to an output. The output is then computed as a weighted sum of the values. \citet{vaswani2017attention} call the attention mechanism they developed ``scaled dot-product attention". The input consists of query and key vectors of dimension $d_k$ and value vectors of dimension $d_v$. The weights for the values are obtained by applying a softmax to the dot product of the query with all keys divided by $\sqrt{d_k}$. In practice, the component vectors of the attention function are packed together into matrices $Q$, $K$ and $V$ and computed simultaneously as: 

\begin{equation*}
    \mathrm{Attention}(Q,K,V) = \mathrm{softmax}(\frac{QK^T}{d_k})V
\end{equation*}

One of the key innovations of the transformer model, is that rather than performing a single attention function with input vectors of size $d_{model}$ (the dimension of the model's hidden size), the query, key and value vectors are linearly projected $h$ times with different, learned linear projections to their respective dimensions of $d_k$, $d_k$ and $d_v$. The above attention function is then performed over each set of inputs yielding $h$ different $d_v$ value vectors. Finally, these are concatenated into a single value vector of dimension $d_{model}$. The creators of the transformer posit that the parallel representations allow the model to attend to information from different representation subspaces at different positions. Formally, the multi-head attention mechanism can be described as:

\begin{equation*}
    \mathrm{MHA}(Q,K,V) = \mathrm{Concat}(\mathrm{head}_1, ..., \mathrm{head}_j)W^O 
\end{equation*}

\noindent
where 

\begin{equation*}
\mathrm{head}_i = \mathrm{Attention}(QW_i^Q, KW_i^K, VW_i^V) 
\end{equation*}

\noindent
and the projections are learned parameter matrices are:

\begin{equation*}
\begin{split}
W_i^Q \in R^{d_{model}\times d_k} \\
W_i^K \in R^{d_{model}\times d_k}  \\
W_i^V \in R^{d_{model}\times d_v} \\
W_i^O \in R^{hd_v\times d_{model}} \\
\end{split}
\end{equation*}

The output of the MHA sub-layer is then passed to the FFN sub-layer described in the next section.

\subsubsection{Feed-Forward Network Sub-Layer}

Each layer of the transformer contains a fully connected feed forward network (FFN) sub-layer. The shallow network consists of two linear transformations around a ReLU activation. Formally, this is expressed as:

\begin{equation*}
    \mathrm{FFN}(x) = \mathrm{max}(0, xW_1 + b_1)W_2 + b_2
\end{equation*}

$W_1$ projects the input from the model's hidden size $d_{model}$ to an intermediate size $d_{ff}$ and $W_2$ projects the transformed input back to $d_{model}$. $b_1$ and $b_2$ are bias terms. The dimensions of each network are identical, however the parameters of the learned projection matrices $W_1$ and $W_2$ differ from layer to layer. The original transformer had $N=6$ layers with dimensions $d_{model} = 512$ and $d_{ff} = 4*d_{model} = 2048$. Because the attention mechanism has no inherent order, the original transformer and most subsequent models add a positional encoding to the input embeddings so the model can make use of the sequence information.

\begin{figure}[]
\centering
\includegraphics[width=4cm, height=8cm]{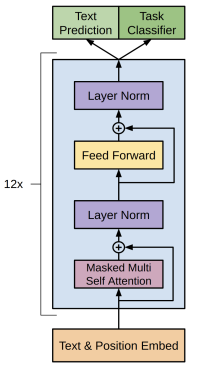}
\caption{Diagram of the GPT model. From \citet{radford2018improving}}
\label{fig:gpt}
\end{figure}

\subsection{Pretrained Transformers for Natural Language Processing}
\label{subsec:transformers-for-nlp}

After their introduction in \citet{vaswani2017attention}, transformer models were quickly adapted to the task of transfer learning for NLP tasks. The GPT (Generative Pretrained Transformer) set new state-of-the-art performance on a variety of tasks upon its introduction by \citet{radford2018improving}. The GPT model consists of a large stack of transformer decoder blocks. The model is pretrained on the BooksCorpus dataset \cite{zhu2015aligning}, a large, unlabelled corpus of text using a standard language modeling objective. That is, given an unsupervised corpus of tokens $U = \{u_1, ..., u_n\}$ the model maximizes the likelihood $L = P(u_i | u_{i-k}, ..., u_{i-1}; \Theta)$ where $k$ is the context size and $\Theta$ are the network parameters. Less formally, the model must predict the next token, given the $k$ previous tokens. After pretraining for 100 epochs, the model can be transferred to a supervised NLP task by replacing the output layer of the network and training for a few epochs. This process is known as fine-tuning.

Shortly after GPT, BERT (Bidirectional Encoder Representations from Transformers) was introduced in \citet{devlin2018bert}, and is now the most widely used NLP model available. In contrast to the generative GPT model, BERT consists of a stack of transformer encoder blocks. Crucially, BERT was the first transformer model pretrained with the masked language modeling objective (MLM). In MLM, the model is presented with a sequence of text tokens, some percentage of which are replaced with a special "[MASK]" token, and must predict the masked tokens given the unmasked tokens. Unlike the standard language objective, the prediction is conditioned on words that come before and after the token being predicted. After pretraining the model can then be fine-tuned to downstream tasks with minimal changes to the model parameters and architecture. 

Pretrained transformers have mostly displaced recurrent neural networks as the standard for NLP tasks. Though the pretraining tasks and domain are often quite different than the down stream tasks they are applied to, they generally outperform task specific deep models. The GPT model is now in its fourth iteration, GPT-4 \cite{openai2023gpt4}. By augmenting traditional language modeling with reinforcement learning, it is capable of reliably producing human quality text on demand. The functionality and performance of BERT have been extended by a variety of models \cite{liu2019roberta, lan2019albert}, and compressed to improve inference time \cite{sanh2019distilbert}.  Inspired by the extraordinary success of using transformers for NLP, researchers have recently begun to adapt transformers to the CV domain. We will close this section by describing some of these efforts.

\subsection{Pretrained Transformers for Computer Vision}
\label{subsec:transformers-for-cv}
The ViT (Vision Transformer) model was introduced in \citet{dosovitskiy2020image}. As the name suggests, it is a transformer based model and its creators closely follow the design and dimensions of the BERT model. In place of textual tokens, features are created by breaking an image into a sequence of $P\times P$ patches, where $P$ is the patch size, flattening the 2-D image patches and then linearly projecting them to match the transformer's embedding size. It is pretrained using supervised image classification on a labelled dataset. The model's creators show that ViT can match or exceed state-of-the-art CNN-based networks on common downstream image classification benchmarks when it is provided with sufficient data. One notable drawback of ViT is that it requires more pretraining data to achieve said results than CNN-based networks. The model's creators surmise that this is because transformers lack the image-friendly inductive biases of CNNs, such as locality, two-dimensional neighborhood structure and translation equivariance. 

\begin{figure}[]
\includegraphics[width=8cm, height=5cm]{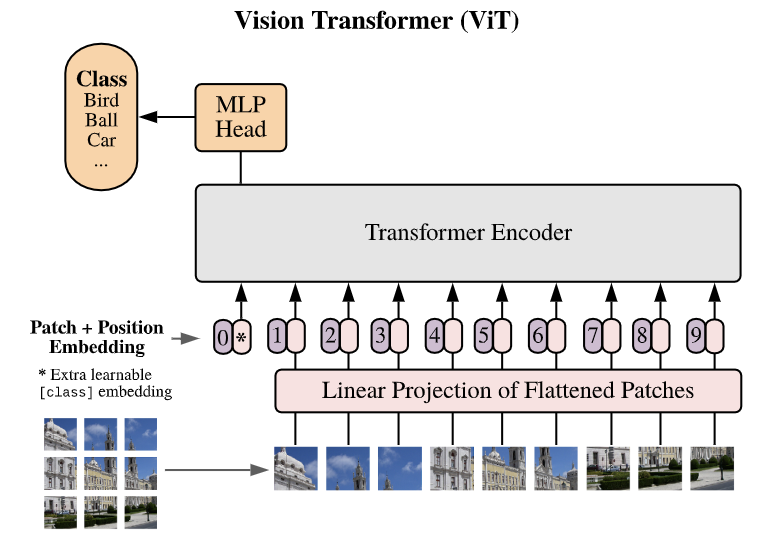}
\caption{Diagram of ViT (VisionTranformer) for computer vision tasks. From \citet{dosovitskiy2020image}}
\label{fig:vit}
\end{figure}

One possible solution to the problems posed by the large pretraining data requirements of ViT was proposed in the BEIT model \cite{bao2021beit}. BEIT adopts a similar architecture and embedding strategy as ViT, but introduces a novel pretraining task, masked image modeling. The masked image modeling task is much likes BERT's masked language  modeling objective. Image patch representations are first "tokenized" to a discrete representation using an auto-encoder.  The designated, abstract category of the masked image patch is then predicted by the  model. The creators show that it can perform on par with ViT using substantially less pretraining data. Notably, the masked image modeling task is very similar to the masked region modeling objective used by some VL transformer models that is described in Section~\ref{sec:pretraining}. 

\begin{figure*}[]
\centering
\includegraphics[width=10cm, height=3cm]{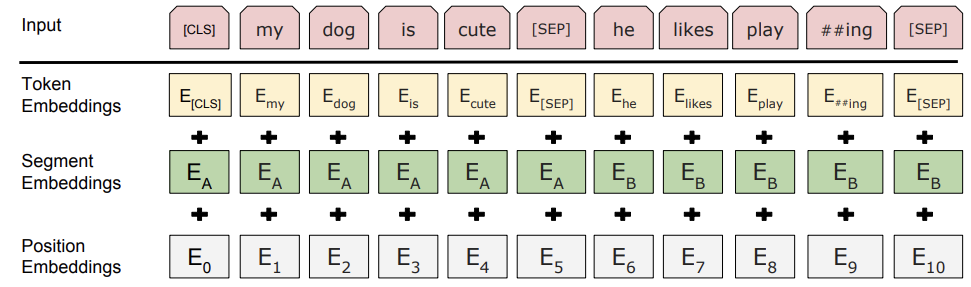}
\caption{Textual Embedding Scheme from BERT. Note that in VL transformers token embeddings are a mixture of visual and text tokens and segment embeddings will often denote whether a token is image or text.  From \citet{devlin2018bert}}
\label{fig:bert-embedding}
\end{figure*}
 
Other approaches to improving the transformer's performance in vision tasks take their inspiration from convolutional neural networks. CoAtNet \cite{Dai2021-ps} employs depthwise convolution layers that combine the data inductive biases of CNNs with the model capacity of transformers. Swin \cite{Liu2021-vy} and CSwin \cite{Dong2021-rj} are transformer based vision models known as hierarchial transformers are another approach. These models perform self-attention over a shifting two dimensional segment of the input image. This mechanism creates a bias toward learning two dimensional relationships and it allows hierarchical transformers to learn with less data than the vanilla vision transformer.

In this section, we have covered sufficient background to fully explore our main topic. The remainder of the paper will be devoted to describing pretrained transformer VL models, starting with the strategies they use to jointly embed visual and textual features.

\section{Embedding Strategies}
\label{sec:embedding}

In this section we discuss how VL transformer models encode their textual and visual embeddings into the model's feature space. Formally, textual and visual input must be encoded into a sequence of textual tokens $\{t_1, .... t_T\}$ and a sequence of visual features $\{v_1, ..., v_V\}$ where each sequence element is a numerical vector. Virtually all of the models that we reviewed for this paper adopt the same embedding strategy for textual representations, described in detail in the subsection immediately below. The strategy for representing images however, varies significantly and represents one of the key differences in pretrained VL models and we will discuss the subject in detail in the following section.

\subsection{Textual Embeddings}
\label{subsec:textual-embeddings}

Most of the VL models we discuss use the textual embedding strategy of BERT \cite{devlin2018bert}. The input text is first tokenized using the WordPiece algorithm \cite{wu2016google}. Each sequence begins with a special classification token "[CLS]" and sequences of text are separated using the special "[SEP]" token. Finally, learned embeddings representing a token's position within the sequence and which segment of text it belongs to are added to produce the token's input representation. The process is summarized visually in Figure~\ref{fig:bert-embedding}. Some models such as CLIP \cite{radford2021learning}, UNIMO \cite{Li2020-up}, OFA \cite{Wang2022-gx} and METER \cite{dou2022empirical} to name a few, use the BPE encoding scheme \cite{sennrich2015neural} as opposed to WordPiece encoding. Other models, BEiT-3 \cite{wang2023image}, VL-T5 \cite{cho2021unifying} and Flamingo for example, make use of the SentencePiece encoding described in \citet{kudo2018sentencepiece}. The essential embedding strategies of these models however, are largely the same.

\subsection{Visual Embeddings}
\label{subsec:visual-embeddings}

\subsubsection{Region Features}
\label{subsubsec:region}

Among the most common approach to creating visual embeddings for VL transformers has been to use region based features produced by an out of the box object detection network. Object detection networks segment images into rectangular regions containing discrete visual objects and assign each region with an appropriate label for the object it contains. Fast R-CNN \cite{girshick2015fast} and YOLO \cite{redmon2016you} are popular examples. UNITER \cite{Chen2019-sh}, ViLBERT \cite{Lu2019-bb}, VL-BERT \cite{Su2019-pf}, VisualBERT \cite{li2019visualbert}, Oscar \cite{li2020oscar}, VinVL \cite{zhang2021vinvl},  LXMERT \cite{Tan2019-yi}, VL-T5 \cite{cho2021unifying}, Unicoder-VL \cite{Li2020-ea} and UNIMO all make use of region based features. Because the attention mechanism used by transformers is inherently unordered, an embedding representing position of the region within the image is usually included added to the feature embedding. 

As an illustrative example, consider the embeddings of VL-BERT. VL-BERT extracts region features from image input using a Fast R-CNN object detector \cite{girshick2015fast}. A visual feature embedding is taken as the last hidden state vector prior to the final output layer for each Region of Interest (RoI). Information regarding the position of the bounding box enclosing the region is embedded into a high dimensional space and included with each RoI feature. The concatenated visual feature embedding is then projected to the matching the dimension of the textual features and concatenated with the VL transformer's textual token embeddings.  Tokens representing image regions are marked with a special "[IMG]" token, textual and other special tokens receive an RoI representation of the entire image. Figure 3.1 gives a visual summary of the model's embedding strategy.

\begin{figure*}[]
\centering
\includegraphics[width=18cm, height=6cm]{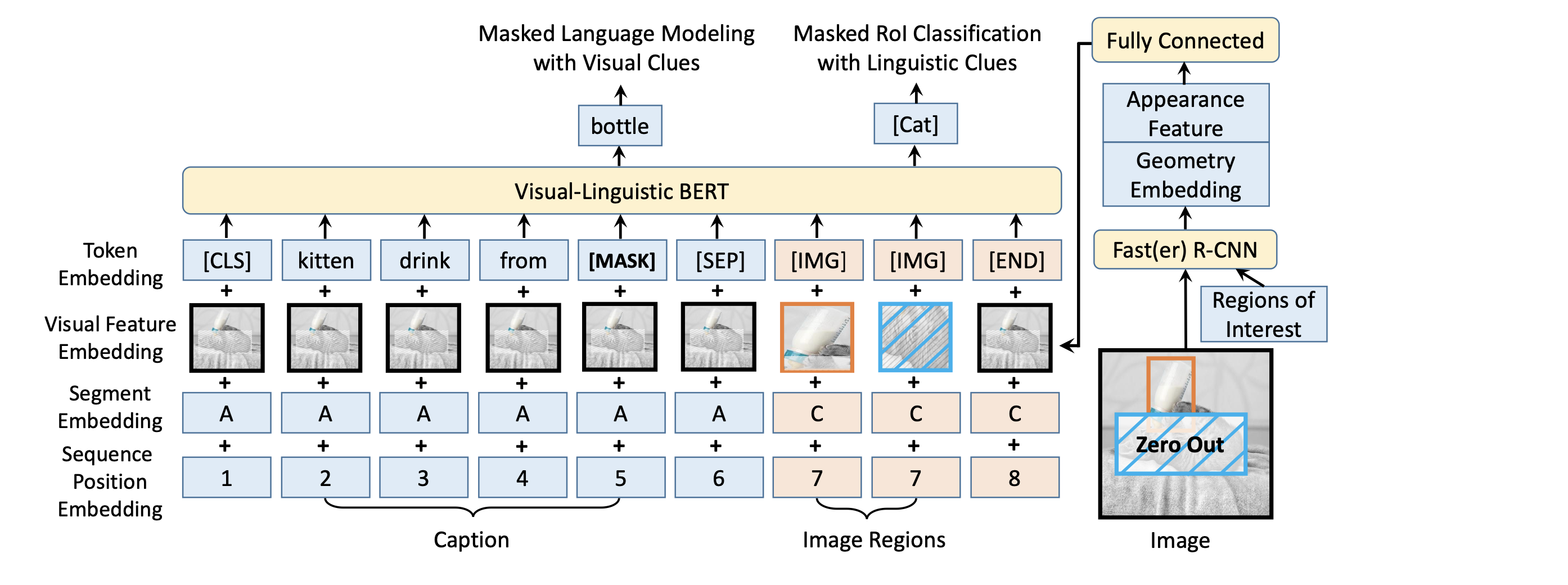}
\caption{Visual embedding scheme using region features for VL-BERT. From \citet{Su2019-pf}}
\label{fig:vl-bert-embedding}
\end{figure*}

The general complexity of region features makes for significant variations in how different model create and use them. UNITER simply concatenates the textual and visual features and separates them with a special "[SEP]" token. This allows it to dispense with the special "[IMG]" token that VL-BERT uses. ViLBERT and LXMERT create RoI features with CNN-based object detectors but keep textual and visual embeddings separate and use a cross-attention mechanism to allow them to interact. There are two notable drawbacks to using region features. Firstly, they can only identify the objects that object-detection model used to create them was trained on. VL models using them, therefore, can only recognize and describe those same visual categories. Secondly, object detection networks used to generate region features are also computationally expensive, and creating region features represents a serious computational bottleneck for models that use them \cite{kim2021vilt}.

\subsubsection{Grid Features}
\label{subsubsec:grid}

In place of region features, the encoder only models Pixel-BERT \cite{huang2020pixel}, and SOHO \cite{huang2021seeing} and the encoder-decoder model E2E-VLP learn to align text with visual embeddings extracted from the feature grid outputted by a CNN. The dual encoders CLIP, ALIGN \cite{jia2021scaling} and LiT \cite{zhai2022lit} also use grid features. As well as the visually grounded transformers GPV-1 \cite{Gupta2021-kn}, mDETR \cite{kamath2021mdetr}, DQ-DETR \cite{Liu2022-pg}, UniTAB \cite{Yang2022-hf}, Referring Transformer \cite{Li2021-wn} and KD-VLP \cite{Liu2022-qj}. We consider the formal expression of the process for creating the grid features used by E2E-VLP. A raw image input is fed into a CNN encoder. Given an image $I$ with 3 color channels and height $H_0$ and width $W_0$ such that $I \in R^{3\times H_0 \times W_0}$ the CNN encoder will output a feature map $f \in R^{C \times H \times W}$. The E2E-VLP model uses the same dimensions as DETR \cite{Carion2020-hb} with embedding dimension $C=2048$ and spatial dimensions $H=\frac{H_0}{32}$ and $W=\frac{W_0}{32}$. To reduce the embedding dimension a $1 \times 1$ convolution is applied to reduce C to a smaller dimension $d$ such that we obtain a lower resolution map $z \in R^{d \times H \times W}$. Finally, in order to produce a sequence of tokens, the feature map is flattened to a sequence of $H \cdot W$ vectors of dimension $d$. The essential strategy of models using grid features is essentially the same in most important respects. 

This approach to visual embedding removes the theoretical ceiling imposed by the object categories of region features. It also provides a dense set of features for fine-grained visual reasoning. However, this approach still relies on an a pretrained CNN as its visual encoder creating a two step training and inference process. And although the creation of grid features uses less compute than an object detection models, image processing still accounts for most of a model like PixelBERT's inference time \cite{kim2021vilt}. We close this section with the following section that describes a strategy called patch embedding that directly addresses the problem of reducing compute requirements for image preprocessing and obviates the need for CNN-based visual encoder.

\subsubsection{Patch Embeddings}
\label{subsubsec:patch}

The final approach we discuss, patch embedding, was introduced by the Vision Transformer (ViT) \cite{dosovitskiy2020image} and was first adapted for use in VL tasks by the ViLT mdoel introduced in \citet{kim2021vilt}. Since its introduction in ViLT, it has been a popular choice and is used by VLMo \cite{bao2022vlmo}, ALBEF \cite{li2021align}, BEiT-3, BLIP-2 \cite{li2023blip}, CoCa \cite{Yu2022-zn}, SimVLM \cite{Wang2021-se}, OFA \cite{Wang2022-gx}, METER \cite{dou2022empirical}, mPLUG \cite{li2022mplug}, BridgeTower \cite{Xu2022-ce}, DaVinci \cite{diao2022prefix}, Florence \cite{Yuan2021-yg} and FLAVA \cite{singh2022flava} models. 

\begin{figure*}[]
\centering
\includegraphics[width=15cm, height=6cm]{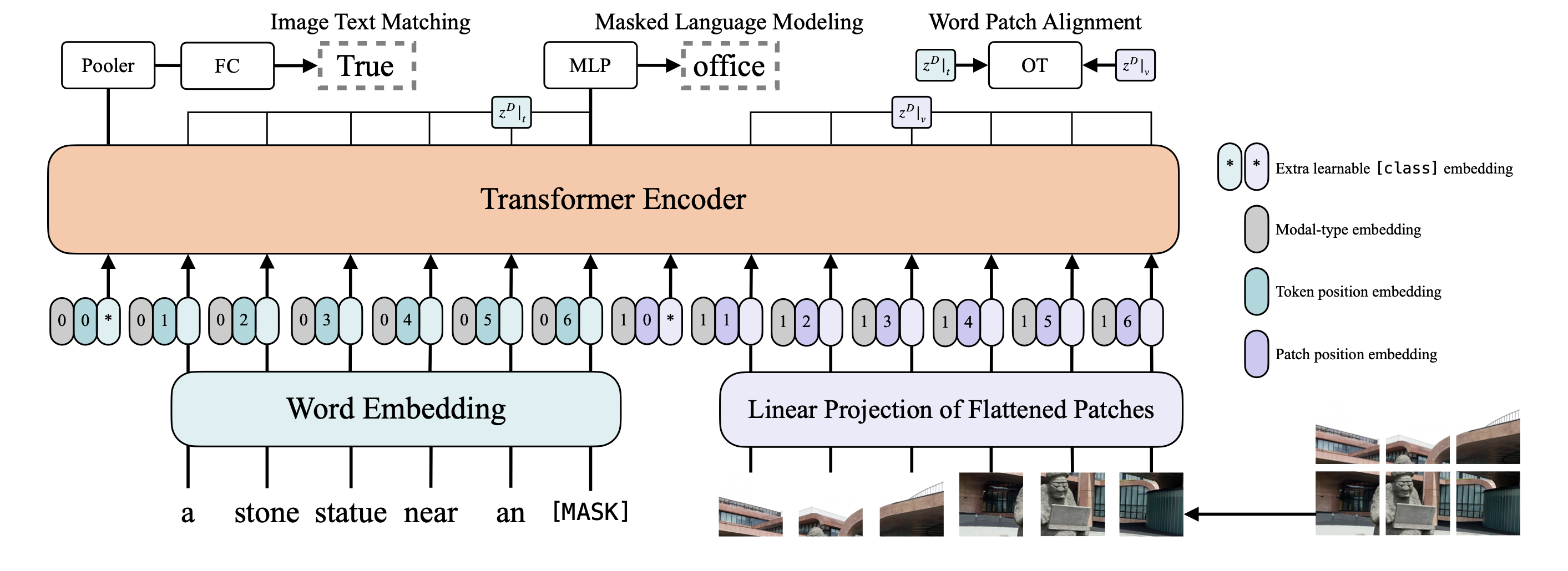}
\caption{Patch Embedding Scheme from ViLT. From \citet{kim2021vilt}}
\label{fig:vilt-patch}
\end{figure*}

Formally, a given image $I \in \mathbf{R}^{3 \times H \times W}$ is sliced into $N = HW/P^2$ patches and flattened into $p \in \mathbf{R}^{N \times (P^2\cdot C)}$ where $(P, P)$ is the patch resolution. A learned linear projection is then used to project each feature to the embedding dimension. The image embedding is obtained by summing the patch projection with learned position and type embeddings. Finally, the input features are formed by concatenating textual and visual embeddings for input into the transformer. Using a patch size of $P=32$, the ViLT model uses only a of fraction of the compute for image processing than the models previously discussed. Figure~\ref{fig:vilt-patch} shows an illustration of the patch embedding process for ViLT provided by \citet{kim2021vilt}. 

Most of the models use ViT \cite{dosovitskiy2020image} to process images into patch features. As a result, they follow a similar approach to patch embedding. Three notable exceptions are OmniVL, OFA and SimVLM, which use a CNN architecture to extract image patches. The authors argue that these patch embedding are superior to those obtained from the simple linear projection used by ViT. 

\section{Model Architecture}
\label{sec:architecture}

Independent of the embedding strategy employed, the model architecture of VL models must allow features associated with the textual and vision modalities to interact in some fashion. In this section we describe the different model designs used by pretrained VL transformers to jointly represent vision and language. In the broadest sense, pretrained VL models can be classified by whether this interaction is achieved through a shallow interaction, such as a dot product, or whether interaction occurs within the deep learning model itself. Among models using a deep interaction, architectures employ either a single-tower encoder, a dual-tower encoder or an encoder-decoder design. Following \citet{bao2022vlmo}, we refer to models using shallow interaction as dual encoders. 
These architectures are described in the subsection immediately below with notable examples from available VL models.

\subsection{Dual Encoders}
\label{subsec:dual}

Dual encoders model visual and textual representations separately and the modalities do not interact within the deep learning model. Instead, the output of the visual and textual modules interact through a simple mechanism, usually a cosine similarity. Notable examples of dual encoder models are the CLIP model from OpenAI and ALIGN from Google Research, Florence and LiT. 

Consider the CLIP model, here text is encoded using a 12 layer transformer with BPE encoding \cite{sennrich2015neural}. Images are encoded with either a ResNet50 \cite{He2015-et} or Vision Transformer \cite{dosovitskiy2020image} architecture. The output embeddings of each model are then compared via cosine similarity, which for normalized features reduces to a vector dot product. The shallow interaction scheme allows for a simple pretraining process that scales well to very large datasets. Given a large number of (image, text pairs), training minimizes cosine similarity between correct pairs and maximizes it between incorrect pairs, a process called contrastive learning which is described in detail in Section~\ref{subsec:contrastive}. 

The ALIGN model follows a strategy very similar to CLIP, using an EfficientNet \citet{} module as an image encoder and BERT as a text encoder. Like CLIP the encoder modules interact via a contrastive loss on their outputs. Crucially, ALIGN massively scales up the data set used for training. Its creators collected 1.8 billion image and alt-text pairs from the internet and performed only minimal post-processing steps in favor of scale. The large, and noisy, dataset allows ALIGN to surpass CLIP on several important benchmarks. Though much like CLIP and ALIGN in terms of architecture, LiT takes the novel approach of using a pretrained image encoder whose weights are locked during its contrastive training. The authors show through ablation studies that this approach has a variety of advantages and it performs better than CLIP and ALIGN several image retrieval benchmarks.

\begin{figure*}[]
\centering
\includegraphics[width=15cm, height=6cm]{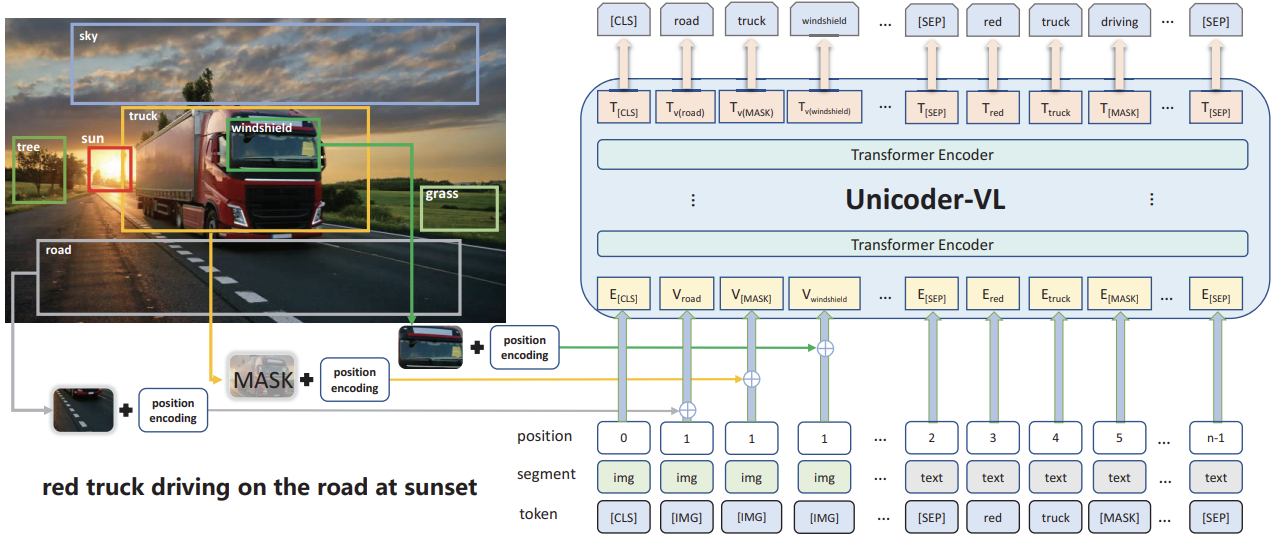}
\caption{The single-tower fusion architecture used by Unicoder-VL. From \citet{Li2020-ea} }
\label{fig:unicoder-single-tower}
\end{figure*}

The Florence model extends the dual encoder to a much greater variety of downstream tasks and capabilities. The base model consists of a text encoder transformer, RoBERTa \cite{liu2019roberta} and a hierarchial vision transformer, CSwin \cite{Dong2021-rj} as the visual encoder trained with contrastive learning. Florence, however, can be equipped with a large number of task-specific heads that allow it perform a wider variety of tasks than the other dual encoder models, including inference on video tasks. 

Despite their relative simplicity, dual encoder models like CLIP achieve remarkable results on a variety of tasks \cite{radford2021learning}, in particular on zero shot classification and retrieval tasks.  The fast inference time of dual encoder models make them ideal for image retrieval tasks. However, the creators of CLIP and other researchers \cite{kim2021vilt} have noted that CLIP performs poorly on complex VL classification tasks such as NLVR2 \cite{suhr2018corpus}. The Florence model performs relatively well on the VQA task \cite{Yuan2021-yg} though well below state-of-the-art and it was not tested on a wide array of complex classification tasks. In order to attain state-of-the-art results on such tasks, a deeper interaction between modalities appears to be required. In the next two sections we explore deep interaction mechanisms that fare better on complex VL tasks.

\subsection{Fusion Encoders}
\label{sec:fusion}

\subsubsection{Single-Tower Architecture}
\label{subsec:single-tower}

Following \citet{bugliarello2021multimodal}, we classify fusion encoders into two categories, single-tower and dual-tower architectures. In a single-tower architecture, a single transformer encoder operates on a concatenation of visual and textual input representations. Since both the visual and textural tokens are embedded into a single input, the single transformer stack allows for unconstrained modality interaction modeling. It also has the distinct advantage of requiring fewer parameters than the more complex two-tower architecture.

Of the two encoder types, single stream architectures have been the most common approach thus far. ViLT, VL-BERT, UNITER, OSCAR, SOHO, UNIMO, PixelBERT, Unicoder-VL and VisualBERT all use a single stream architectures. We will use Unicoder-VL as an illustrative example. The region features and textual input embeddings described in Section~\ref{sec:embedding} are fed into a transformer encoder with the same dimensions as the BERT model \cite{devlin2018bert}. The architecture is depicted in Figure~\ref{fig:unicoder-single-tower}

Though single-tower models differ along other dimensions, embedding strategy, pretraining tasks and data used, etc., the essential architecture is much the same for all of single-tower models. Many of them, as their names suggest, are variations on the BERT NLP model and some, such as VisualBERT and VL-BERT are initialized with the pretrained weights from BERT. A notable feature of the aforementioned ViLT, is that  it is initialized with the pretrained weights from ViT \cite{dosovitskiy2020image} instead of BERT. 

\begin{figure*}[t]
\includegraphics[width=15cm, height=4cm]{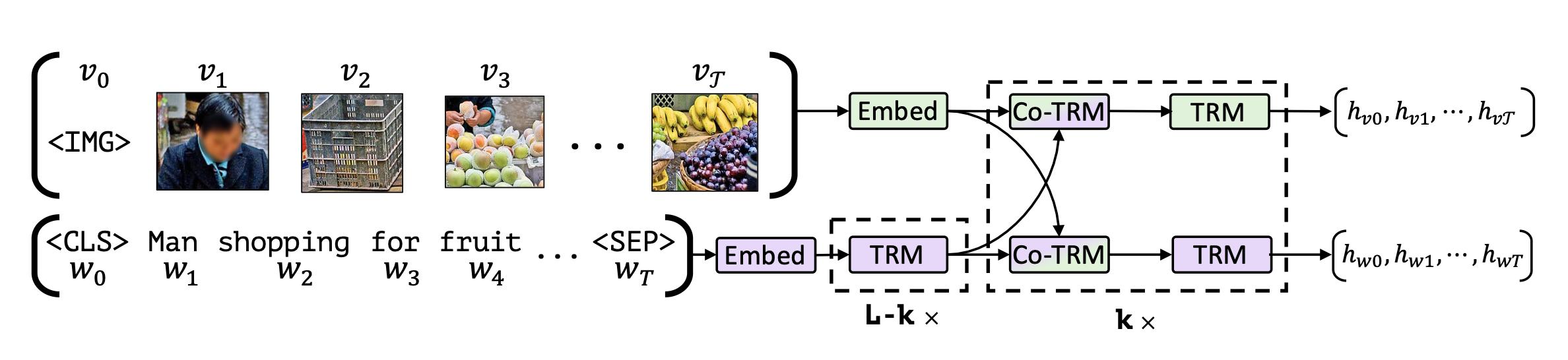}
\caption{Depiction of cross-attentions in the two-tower fusion encoder architecture of ViLBERT. From \citet{Lu2019-bb}}
\label{fig:vilbert-two-tower}
\end{figure*}

\subsubsection{Two-Tower Architecture}
\label{subsec:two-tower}

Rather than operating on a concatenation of visual and textual inputs, two-tower architectures encode each modality in separate transformer stacks. Interaction is then achieved through a cross attention mechanism. ViLBERT, LXMERT, METER and BridgeTower are examples of dual-tower models. In ViLBERT, for example, separate image and text embeddings are passed into parallel sequences of BERT-style transformer blocks and co-attention transformer layers. Co-attention layers take intermediate visual and textual representations $H^{(i)}_V$ and $H^{(i)}_T$ and compute query, key and value matrices as in a standard transformer block. The keys and values for each modality are then passed as input into the other modality's multi-headed attention block creating a multi-modal feature. The rest of the layer proceeds as in a traditional transformer block, resulting in a multi-modal feature. ViLBERT's architecture is depicted in Figure~\ref{fig:vilbert-two-tower} LXMERT's architecture and general approach are quite similar to that of ViLBERT. One notable difference between the models is that while ViLBERT's text module is intialized with the weights from BERT-base, all of LXMERT's parameters are trained from scratch. 

METER, though broadly similar to LXMERT and ViLBERT,  differs in a number of key ways. Firstly, it creators conducted a broad architecture analysis and attempted a number of different text and image encoders. Their default model, consisted of a pretrained RoBERTa text encoder and a pretrained CLIP-ViI-224-32 \cite{radford2021learning} as the image encoder. These architectural innovations allow METER to outperform the previous models that made use of the patch embeddings mentioned in Section~\ref{sec:embedding}. The BridgeTower model is similar in most respects to METER, however it contains novel features that its creators call bridge connections. Here representations from the top 6 layers of each unimodal encoder are connected to each of the layers in the multi-modal encoder prior to the cross-attention mechanism. This is done on the theory that since different layers of transformers encode different types of information, that these bridge connections will enrich the model's multi-modal representations. And indeed, BridgeTower is able to outperform METER on several downstream benchmarks despite being almost identical in other respects. 

\subsection{Combination Encoders}
\label{subsec:combo-encoder}

\begin{figure*}[t]
\includegraphics[width=15cm, height=12cm]{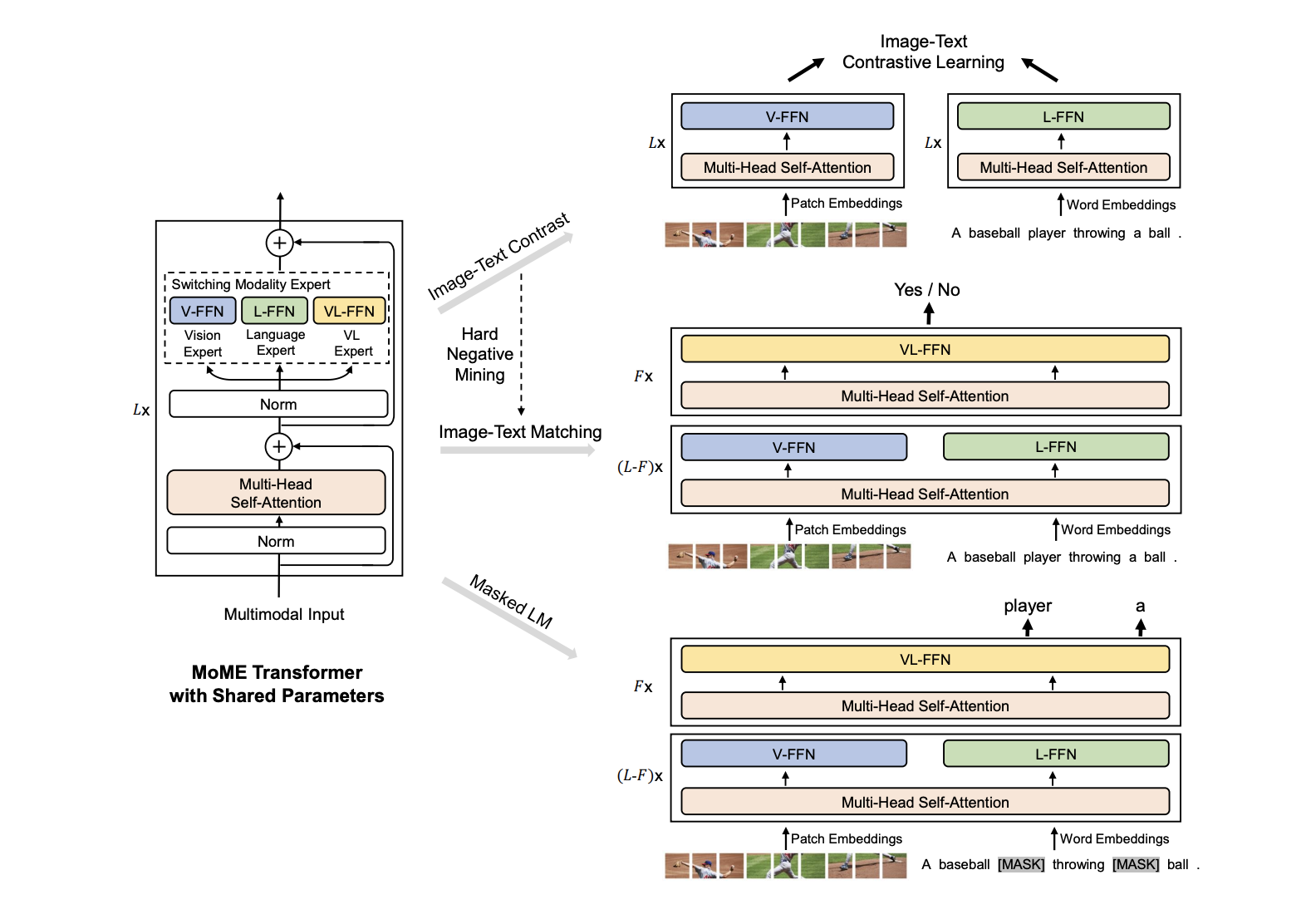}
\caption{Combination Dual-Fusion Encoder Architecture for the VLMo model. From \citet{bao2022vlmo}}
\label{fig:vlmo-combo-architecture}
\end{figure*}

Several recently introduced VL modes,VLMo, ALBEF, BEiT-3 , FLAVA and $\text{X}^2$-VLM have attempted to leverage the strengths of both dual encoders and fusion encoders in a single model. These models contain separate visual and textual encoders at the base of the model.  The outputs of the text encoder and an image encoder are aligned using cosine similarity before being fed into a fusion encoder module of some kind. The VLMo model, for instance, combines the two interaction schema via a transformer block the model's creators call the mixture of modality experts. Here the FFN of a transformer encoder block is replaced with pool of modality specific expert models. Specifically the model can direct input to three different modality experts: a vision expert (V-FFN), a language expert (L-FFN) and a vision language expert (VL-FFN). If the model is given purely language or text inputs, it can encode them with the appropriate expert module. When given an image-text pair, it can encode each component with the correct modality expert at lower layers before using the vision language expert module in the top layers. This process is visually summarised in Figure~\ref{fig:vlmo-combo-architecture}. 

The architecture and approach of BEiT-3, is very similar to VLMo, however it is massively scaled up, leading to increased downstream performance on a variety of tasks. FLAVA uses a slightly less complex architecture but the idea remains the same. In FLAVA, the output each modality encoder can be passed to a task head for classification for uni-modal tasks or be passed to a multi-modal encoder after a contrastive loss for vision language tasks. 

The ALBEF model takes a slightly different approach in combining dual and fusion encoder. Here the respective modalities are encoded and then aligned using the cosine similarity and contrastive loss, much like the CLIP or ALIGN models. After this step, the representations are then fed into a 6-layer fusion encoder that uses cross-attention. Though this approach yields a model less flexible than that of a model like VLMo, its novel approach offers the possibility of a deeper, more comprehensive vision language interaction. $X^2$-VLM also consists of three separate modules, a vision encoder, a text encoder and a fusion encoder with features visual and textual features aligned using cosine similarity. Significantly, this model is larger in scale, is trained on more data and can also perform vision language tasks using video as input. 

\subsection{Encoder Decoder Models}
\label{subsec:encoder-decoder}

Following the architecture of the original transformer, some VL models opt for a design consisting of at least one encoder stack and a decoder stack. The VL-T5, OFA, OmniVL, PaLI, E2E-VLP, SimVLM, mPLUG, Flamingo, OmniVL and CoCa models all make use of this architecture. Additionally, the models mDETR, DQ-DETR, UniTAB, KD-VLP and Referring Transformer that all specialize in visual grounding tasks use some variation of the encoder-decoder  design. This model architectures is versatile in general and allows models using them to successfully perform a wide range of functions, including generative tasks such as image captioning. 

As an illustrative example we consider VL-T5. It consists of a single multi-modal encoder stack that accepts a sequence of word-piece tokens and image region features. The encoder produces an intermediate output $h = \{h_1^t, ... , h_{|t|}^, h_1^v, ... , h_{|v|}^v\}$ where $h_i^t$ and $h_i^v$ respectively represent the intermediate state vector for textual and visual representations. The decoder stack then iteratively attends to its own previously generated output via self attention and the encoder's intermediate outputs via cross-attention. Using this procedure the decoder computes the probability of future text token j, formally expressed as $P_{\theta}(y_j|y_{<j},h)$.

Though the general approach in all of the aforementioned architectures is basically similar, there is a very wide range of design distinctions. mPLUG for instance, uses a two-tower encoder of the type discussed just above. Notably, its encoder omits cross attention on some of the layers for its visual component resulting in a more efficient model. Flamingo, uses a novel module called a Perceiver Resampler to convert the output of a pretrained visual encoder to a fixed number of visual features. The visual features are then cross-attended by a text-decoder with alternating blocks of a pretrained large language model and novel cross-attention layers. These features allow Flamingo to handle an arbitrary mix of image and text inputs and generate text in an open ended way. 

OmniVL makes use of two encoder stacks a visual encoder, that can handle both images and video as well as a text transform based on BERT. A novel feature of this model is that it makes use of two decoder stacks, one decoder for vision language alignment and understanding tasks and a separate decoder for generative tasks. CoCa also uses two decoder networks, though, opposed to OmniVL, the outputs of its image encoder and its text decoder are then input to a multi-modal decoder for generative tasks. 

The models mDETR, DQ-DETR, UniTAB and Reffering Transformers, designed for visual grounding tasks, also use encoder-decoder architectures. As their names suggest, mDETR and DQ-DETR are derived from an transformer based object detection model called DETR \cite{Carion2020-hb}. Both of these models process image input with a CNN backbone and text using a transformer encoder. Their output is concatenated and fed to a transformer encoder. The decoder module has a series of learned embeddings called \emph{queries}. The decoder attends to the encoder output and the query embeddings. Finally, the decoder output is sent to a feed-forward network for classification. Though complex, this process provides the model with the type of fine-grained information about visual objects required for visual grounding tasks. The Referring Transformer also makes use of \emph{query} embeddings to extract object information, however it also contains a seperate query encoder module to further process visual object information. The last of these models, UniTAB, uses a fairly generic encoder-decoder architecture.  

Three models, DaVinci and OFA can perform a truly impressive range of tasks and do so without the significant changes in architecture often required to adapt pretrained models to various tasks. The decoder module in both models can generate both image and textual output, meaning that these models can perform vision language understanding tasks such as VQA \cite{Antol2015-cs} as well as generative tasks like image-caption generation. Notably, both of these models can also perform text-to-image generation tasks, like models such as DALLE \cite{Ramesh2021-zk} perform, as well as unimodal tasks such as GLUE \cite{Wang2018-fr} and image classification on datasets such as ImageNet \cite{russakovsky2015imagenet}.

\section{Pretraining Tasks}
\label{sec:pretraining}

This section is devoted to discussing the pretraining tasks used by the various VL tranformers. Pretraining is a key element of these models success and we will devote a significant amount of space to describing these methods. Almost all of the fusion and combination encoder models make use of masked language modeling and image text matching, both of which are extensions of the pretraining objectives used in the BERT NLP model \cite{devlin2018bert}. Below, we describe these tasks in detail along with several additional objectives found in the relevant literature.

\subsection{Masked Language Modeling}
\label{subsec:mlm}

Some variation of the masked language modeling objective introduced in \citet{devlin2018bert} and described in section 2.2, is used by all of the fusion encoder and combination encoder models that we reviewed. Several of the encoder-decoder also make use of variations on this task such as VL-T5. The key difference between this task and masked language modeling in pure NLP is that in the VL setting, the models have access to corresponding visual tokens in addition to unmasked text representations. We refer to PixelBERT as a representative example. Formally, it minimizes the negative log-likelihood:
 \begin{equation*}
     L_{MLM}(\theta) = -E_{(t,v) \sim D}\log P_{\theta}(t_m|\mathbf{t}_{\backslash m}, \mathbf{v})
 \end{equation*}
Where $t_m$ is the masked token being predicted, $\theta$ are the model's parameters, $\mathbf{t}_{\backslash m}$, the unmasked text tokens, and $\mathbf{v}$ are the corresponding image features.  Each pair $(\mathbf{t}$, $\mathbf{v})$ is sampled from the whole the training set $D$. As in BERT, 15\% of test tokens are selected at random and replaced with a special "[MASK]" token.

Though all of the fusion encoders make use of MLM, there are some minor differences in approach worth noting. ALBEF \cite{li2021align} for instance minimizes a cross-entropy loss function rather than a log-likelihood. ViLT \cite{kim2021vilt}, masks only whole word units rather than the subword units that most of these models use in tokenization. This is done on the theory that visual clues are more likely to be associated with whole words; e.g. "giraffe" will have more visual meaning than "gi\#\#" or "\#\#raffe". UNIMO masks only contiguous spans of complete words as is done in SpanBERT \cite{joshi2020spanbert}. Most models using this task, such as SOHO, follow \citet{devlin2018bert} and mask 15 percent of text tokens. Some models deviate from this practice however; BEiT-3 for example masks 40 perent of the text tokens from the image-text pairs in its pretraining dataset.


\subsection{Masked Image Modeling}
\label{subsec:mim}

The masked image modeling task in an extension of the masked language modeling objective described above. In this setting, however, the objects being modeled are image features rather than text tokens. ViLBERT, UNITER, VL-BERT, LXMERT, Unicoder-VL, BEiT-3, UNIMO, SOHO and FLAVA all make use of this objective in one way or another. We consider ViLBERT  as an example. In ViLBERT, 15\% of both text and image regions are randomly selected and masked. Rather than being replaced by a special "[MASK]" token as textual features are, the image features are set to 0. The model then predicts a distribution over the category labels for its region features. Training minimizes the KL divergence between this distribution and the output distribution for the region from the same model R-CNN model it uses for feature extraction. LXMERT, Unicoder-VL, UNITER, UNIMO and VL-BERT perform the masked image modeling classification task in a similar manner. 

Models that do not use region features must approach this task differently since they do not have labels naturally associated with each visual feature. SOHO uses grid features and it uses the index from its visual dictionary module that it assigns to each feature as the label for classification.  BEiT-3 and FLAVA, which uses patch features,  use a discrete variational autoencoder \cite{Ramesh2021-zk} to assign each image patch feature a fixed set of discrete image codes using the process described in \citet{peng2022beit}. The model must then classify each masked patch according to its assigned visual token.

There are a couple of other noteworthy distinctions in the way this task is deployed among the various models. UNITER and LXMERT for example, extend the masked region modeling task to include region-feature regression. Here, the model is tasked with predicting the actual feature values of a masked region using a loss function such as squared error.  Another key difference is that models such as UNITER will only mask elements from one modality at a time; i.e. during masked region modeling all text features are be unmasked and available to the model vice-versa.  VilBERT and LXMERT, on the other hand mask a given percentage of all tokens at anytime regardless of their modality.

\subsection{Image-Text Matching}
\label{subsec:itm}

Most of the encoder only models and some of the encoder-decoder models, including OFA and mPLUG, make use of the binary classification image-text matching task. Here the model, given an image-text pair must determine if the text in fact describes the image. We refer to UNITER as an example. Given a text-image pair $(\mathbf{t}, \mathbf{v})$, the model output is fed into a fully connected layer and a sigmoid function to produce an output score $S_{\theta}(\mathbf{t}, \mathbf{v})$ between 0 and 1. The model then minimizes loss function
    $\mathbf{L_{ITM}(\theta)} = -E_{(t,v) \sim D}[y\log S_{\theta}(\mathbf{t}, \mathbf{v}) + (1-y)\log(1 - S_{\theta}(\mathbf{t}, \mathbf{v}))]$
. Where $y \in \{0,1\}$ is the label indicating whether the text and image are correctly paired. The negative pairs are created by replacing the image or text in paired sample with one randomly drawn from the training set $D$. The task is relatively simple and there aren't many variations between its implementations among models. 

\begin{figure}[t]
\includegraphics[width=7cm, height=5cm]{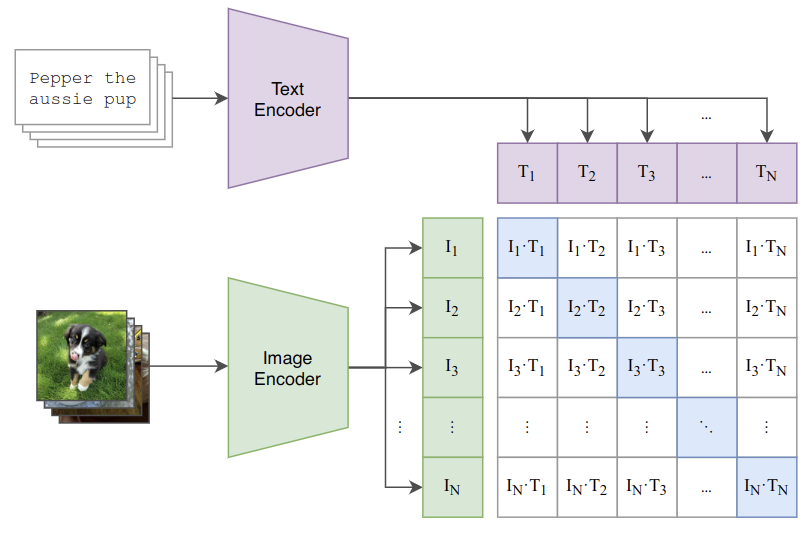}
\caption{Contrastive pretraining from CLIP. From \citet{radford2021learning}}
\label{fig:clip-contrastive}
\end{figure}

\subsection{Contrastive Learning}
\label{subsec:contrastive}

Contrastive learning is the primary pretraining objective employed by dual encoder models such as  CLIP, ALIGN, LIT and Florence and is also used in the combination models ALBEF, VLMo, X2-VLM and FLAVA. Several of the endocder-decoder type models also use some type of contrastive loss, including COCA, mPLUG and Flamingo. We consider the contrastive loss in CLIP as a representative example. Given a batch of $N$ encoded (image, text) pairs, the CLIP model is trained to predict which of the $N^2$ possible (image, text) pairs actually occurred. This is achieved by jointly training an image encoder and a text encoder to minimize the cosine similarity of the $N$ correct pairs and maximize the cosine similarity of the $N^2 - N$ incorrect pairs. The similarity scores are input to a binary cross entropy loss function to be optimized. For normalized values, the cosine similarity reduces to a simple dot product. This process is visually summarized in Figure~\ref{fig:clip-contrastive}. 

Florence uses a slightly modified approach called unified image-text contrastive learning that was introduced in \citet{Yang2022-gd}. The unified approach uses a text hash-table to allow a single text description to be associated with multiple images, a capability not present in the procedure described above. The contrastive loss used in UNIMO is heavily augmented through text rewriting and image/text retrieval to create a large volumes of positive and negative examples. Among the minor distinctions among contrastive loss objectives are approaches to normalizing features. Some models such as CLIP, use the l2 function to normalize features, while others such as VLMo and ALBEF employ a softmax normalization function. A notable feature of contrastive learning is a relatively simple objective function and it scales well to large datasets. 

\subsection{Visual Question Answering}
\label{subsec:vqa}

Some VL transformers use variants of the visual question answering task where the model must choose or generate an answer given as part of their pretraining regime. Some example image-question pairs are displayed in Figure~\ref{fig:vqa} as visual aid. Of the models we reviewed, LXMERT, OFA and VL-T5 use visual question answering. LXMERT treats VQA as a classification task where the model must choose the correct answers from set. VL-T5 and OFA on the other hand leverage their decoder module to simply generate the text given an image, question pair. 

\begin{figure*}[t]
\includegraphics[width=15cm, height=4cm]{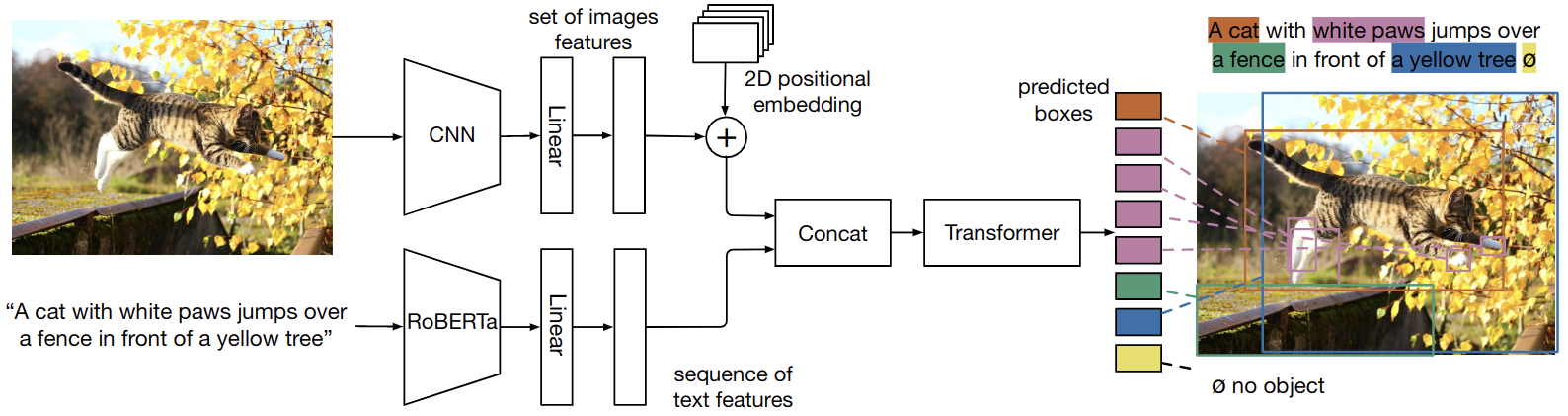}
\caption{Depiction of the visual-grounding tasks used by mDETR for pretraining. From \citet{kamath2021mdetr}}
\label{fig:mdetr}
\end{figure*}

\begin{figure}[]
\includegraphics[width=7cm, height=5cm]{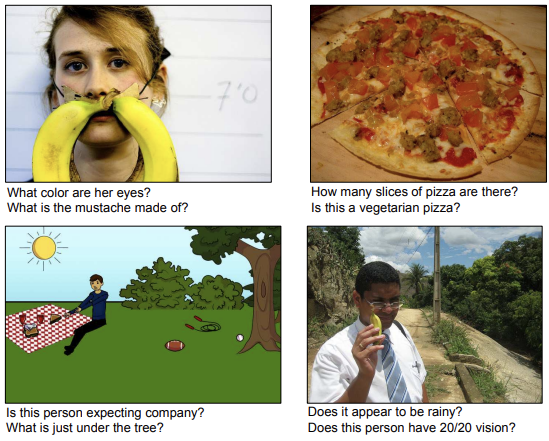}
\caption{Examples from the Visual Question Answering dataset. The VQA task is used by several models for pretraining. From \citet{Antol2015-cs}}
\label{fig:vqa}
\end{figure}

\begin{figure*}[]
\includegraphics[width=15cm, height=6cm]{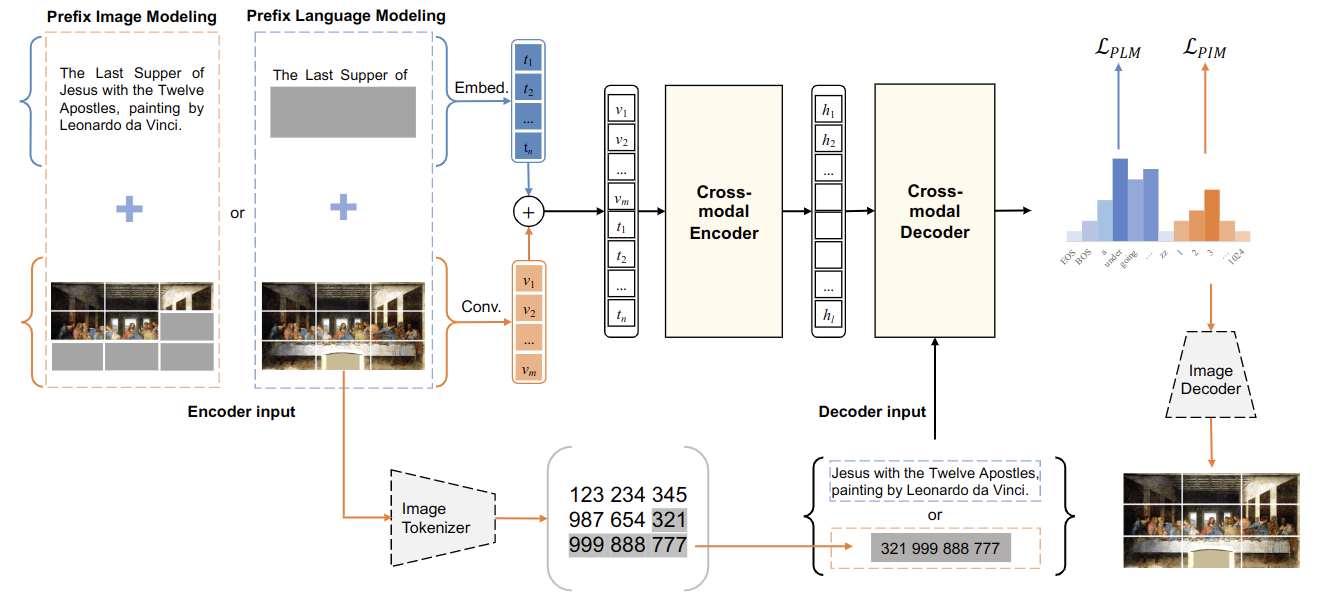}
\caption{Depiction of the prefix modeling task used by DaVinci for pretraining. From \citet{diao2022prefix}}
\label{fig:davinci-prefix}
\end{figure*}

\subsection{Visual Grounding}
\label{subsec:visual-grounding}

Visual grounding tasks are those that connect a specific visual object with the words or phrases that describe them. Reference resolution, identifying which object within an image a given phrase refers to, is an example of visual grounding task. Publicly available datasets such as RefCOCO \cite{kazemzadeh2014referitgame} can be used to benchmark reference resolution. Grounded captioning is the inverse task, where the model is given an image region and must generate a caption that accurately describes it. Visual grounding tasks are of particular interest because they provide direct, fine-grained connections between words and specific visual objects.

Three versatile, general purpose VL models, OFA, X2-VLM and VL-T5 make use visual grounding tasks. OFA uses both reference resolution and grounding captioning as proxy tasks. For reference resolution, it learns to generate a location that specifies the region position $<x1, y1, x2, y2>$ based on the input of the image along with the
instruction "Which region does the text x describe?" For the grounded captioning task OFA learns to generate a description based on the input image and the instruction "What does the region describe? region: $<x1, y1, x2, y2>$". VL-T5 performs visual captioning in essentially the same fashion. VL-T5 performs reference resolution slightly differently, however.  

Additionally, visually grounded transformers all make use of some form of visual grounding task. mDETR for example uses the same technique as DETR \cite{Carion2020-hb} to predict bounding boxes for visual objects within each image. It then assigns each predicted object a probability of being associated with each word in the referring phrase. This process is depicted in Figure~\ref{fig:mdetr}. The other models designed for grounding tasks, such as DQ-DETR, Reffering Transformer and UniTAB use a variety approaches and the interested reader is encouraged to consult their source papers for more detail.

\subsection{Image Captioning}
\label{subsec:image-captioning}

Image captioning requires a model to generate a description of a provided image input. As a proxy task it differs from grounding captioning in that the caption must describe the entire image. Three of the models we reviewed, OFA, E2E-VLP and CoCa, use image captioning as a pretraining objective. In principal, the image captioning task is just the causal language modeling task that uses an image as context to condition the prediction of text tokens. CoCa, for instance formally defines the task as the conditional log likelihood
\begin{equation*}
    L_{cap} = -\sum_{t=1}^T \log{P_{\theta}}(y_t|y_{<t, x})
\end{equation*}
where T is the sequence length, $y_t$ is the token being predicted, $y_{<t}$ are the model's previous predictions and x are the associated image representations.

\subsection{Prefix Language Modeling}
\label{subsec:prefix}

Prefix language modeling can be seen as hybrid between traditional causal language modeling and masked language modeling. Three of the models we reviewed, DaVinci, mPLUG and SimVLM, make use of this training objective. SimVLM, for instance formulates prefix modeling in the following way, a given sequence of tokens is truncated to randomly selected length $T_p$. The sequence of tokens $\mathbf{x}_{<T_p}$ is denoted as the prefix and $\mathbf{x}_{\geq T_p} $ as the suffix. Bidirectional attention is then applied to the prefix sequence and autoregressive language modeling is applied to suffix sequence. In both the SimVLM and mPLUG models, the prefix sequence is chosen in such a way that it always contains all of the image tokens and the suffix consists entirely of text. 

The DaVinci model uses a similar prefix language modeling task and also extends the notion to a prefix image modeling task. Here the model is presented with a prefix consisting of a complete sequence of text tokens and a partial sequence of image tokens. The model must then restore the image tokens in the suffix given the prefix and its previous output. The task is depicted in Figure ~\ref{fig:davinci-prefix}.

\subsection{Other Objectives}
\label{subsec:other-objectives}

The previous sections in this section cover most of the pretraining objectives used by the models reviewed for this paper. However, there a few other noteworthy examples that bear mentioning before we close this section. PixelBERT takes a random sample of its pixel grid features as a form of pretraining regularization. This task of course, can only employed by models like PixelBERT use grid features. Finally, the VLMo model uses a stage-wise pretraining procedure. The VLMo model is a gated mixture of experts network consisting of vision, language and multi-modal sub networks. It first trains the vision expert using BEIT \cite{bao2021beit} style masked imgage modeling while the language and multi-modal weights are frozen. Then it proceeds to training to the language expert with text-only masked language modeling with the vision weights frozen. Finally, the full model is trained with an image-text matching objective. This more or less exhausts the approaches to pretraining used in VL transformers and we have largely summarized most of the meaningful differences among the various VL transformers. Several models, such as UNIMO, employ unimodal training tasks that deal exclusively with language or vision without freezing parts of the model as is done in VLMo. 

\section{Downstream Capabilities}
\label{sec:capability}

In principle, most of the models that we've discussed can be adapted to almost any given VL task with the proper adjustments to model architecture and a fine-tuning regime. Many models however, were explicitly designed and tested for certain VL capabilities. Dual encoders, for instance are especially well suited for alignment tasks such as text to image retrieval. The grounded transformers, e.g. mDETR or Referring Transformer, were trained on and extensively evaluated on visual grounding tasks.  In this section we briefly cover the range of VL tasks that the creators of the models we reviewed either pretrained, evaluated zero-shot or fine-tuned on their models on. In the process, we will take the opportunity to reference some of the major benchmarks used for each type of task. 

\subsection{VL-Alignment}
\label{subsec:alignment}

We define vision language alignment tasks as those tasks such that given a token from one modality, they must correctly relate a set of tokens from the other modality. Retrieval tasks are the canonical examples of alignment task. For example, in text to image retrieval, a model is given an text description and must select and rank a set of matching images. MSCOCO \cite{Lin2014-ju} and Flickr30K \cite{plummer2015flickr30k} are two publicly available datasets that are often used as benchmarks. The vast majority of models we have reviewed have been evaluated on at least one retrieval task. Dual encoder models such as CLIP, ALIGN and LiT however, are designed for VL-alignment tasks. Because image representations for these encoders can be cached, the simple nature of the models allow for quick scoring of large sets of image-text pairs. This makes dual encoders ideal for alignment tasks such as image searches.

\subsection{VL-Understanding}
\label{subsec:vl-understanding}

\begin{figure}[]
\centering
\includegraphics[width=7cm, height=5cm]{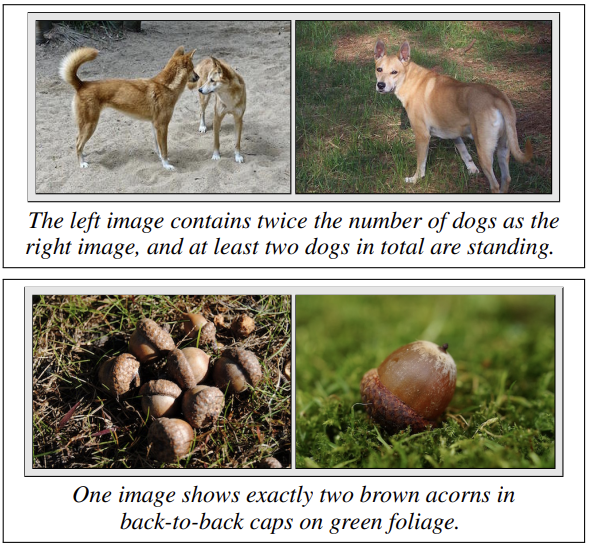}
\caption{Examples from the NLVR2 task. Model must correctly answer true or false. This is a common benchmark for VL-understanding. From \citet{suhr2018corpus}}
\label{fig:nlvr2}
\end{figure}

In vision language understanding tasks a model must correctly classify the relationship between a given image text pair. VL understanding tasks include popular benchmarks such as VQA \cite{Antol2015-cs}, where a model must choose the correct answer to a question about an image, NLVR2 \cite{suhr2018corpus}, where a model must determine if a statement about an image is true or false. Example image-text pairs from the NLVR2 dataset are shown in Figure~\ref{fig:nlvr2}. All of the fusion encoders that we've discussed, both the one-tower and two-tower variations, are fine-tuned and benchmarked on VL-understanding tasks. Encoder-decoder models are also well tested on understanding tasks, though often they recast them as text-generation tasks. OFA, for instance, performs VQA by providing the model with an image and question and allowing it to generate the correct answer. This approach allows some encoder-decoder models to perform multiple types of VL-understanding tasks without the architectural changes encoder only models usually require. With the exception of Florence, most dual encoders aren't well tested on understanding tasks. 

\subsection{VL-Text Generation}
\label{subsec:vl-text-generation}

Vision language text generation tasks are those such as image captioning, where a model is presented with an image or an image-text pair and must generate an appropriate sequence of text describing the image. MSCOCO Captions \cite{Chen2015-my} and nocaps \cite{Agrawal2019-lv} are two common image captioning benchmarks. The models best suited to tasks of this nature are those that contain a decoder module. Encoder-decoder models such as, CoCa, SimVLM, mPLUG, OFA, DaVinci, OmniVL and GiT are all evaluated on text generation tasks. Despite not having a decoder that is optimized for text generation, the encoder-only models BEiT-3, OSCAR, VinVL, X2-VLM and UNIMO were evaluated and perform well on image captioning tasks. None of the visually grounded transformers we reviewed were explicitly evaluated on VL text generation task, though, in principle they could be adapted to the task without too much difficulty.

\subsection{Visual Grounding, Grounded Captioning and Object Detection} 
\label{subsec:gounding}
Visual grounding tasks are those which require a model to match words or phrases to the distinct visual objects they describe. RefCOCO \cite{kazemzadeh2014referitgame} is the premier dataset for benchmarking visual grounding tasks. Grounded captioning, accurately describing a specific object within an image, and visual object detection are related tasks. The visually grounded transformers, mDETR, DQ-DETR, KD-VLP, Referring Transformer and UniTAB were all designed, trained and evaluated specifically for these types of grounded modeling. The encoder-decoder models OFA and VL-T5 are also pretrained and evaluated on grounding related tasks. And though not pretrained on any grounding related tasks, the encoder only models ALBEF, BEiT-3, LXMERT, UNITER, VilBERT, VisualBERT AND X2-VLM and the endcoder-decoder models BLIP-2, E2E-VLP and mPLUG were all evaluated on visual grounding or object detection tasks.

\subsection{Image Generation}
\label{subsec:image-generation}

\begin{figure}[]
\centering
\includegraphics[width=6cm, height=4cm]{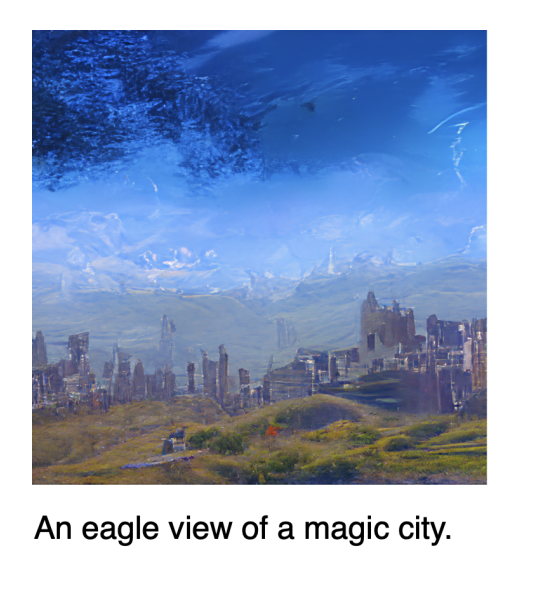}
\caption{An image generated by OFA with the text that prompted it. From \citet{wang2023one}}
\label{fig:ofa-image}
\end{figure}

In text to image generation, such as that performed by DALL-E \cite{Ramesh2021-zk}, a model provided with a text description must output an appropriate image. Most of the models discussed in this paper of geared toward classifying or generating text related to images. Two models however, OFA and DaVinci, are capable of generating images in addition to the other sets of tasks described above. An image generated by OFA with the text that prompted it is displayed in Figure~\ref{fig:ofa-image}.

\begin{figure*}[]
\centering
\includegraphics[width=14cm, height=6cm]{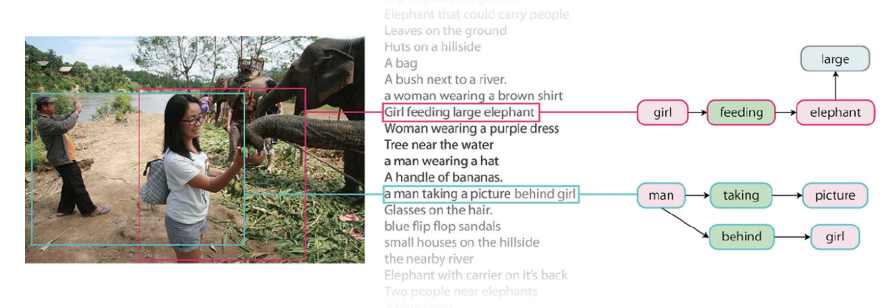}
\caption{Picture from The Visual Genome dataset. Examples of bounding boxes, region descriptions and scene graphs are shown. From \citet{krishna2017visual}}
\label{fig:vg}
\end{figure*}

\subsection{Video Tasks} 
\label{subsec:video}
Though video-based models have not been the focus of this paper, it is worth noting that a number of models can perform video retrieval and understanding tasks in addition to the tasks previously described. Models such as CoCa, Florence, GiT, and Omni-VL were all designed and trained for video tasks and are evaluated on a variety of video benchmarks. 

\subsection{Unimodal Tasks} 
\label{subsec:unimodal}
Several models are also trained to and evaluated on vision-only or text-only tasks. The dual encoder models such as CLIP and ALIGN are easily adapted to pure computer vision tasks such as image classification are all evaluated on such tasks. Many of the flexible encoder-decoder models are also evaluated on image classification. These models include BLIP-2, CoCa, DaVinci, FLAVA, Flamingo, GIT, OFA, OmniVL and PaLi. Surprisingly, from the fusion encoder models that we reviewed, UNIMO is the only one that was specifically trained and evaluated on unimodal tasks. Many fewer models have been evaluated on pure NLP tasks such as the GLUE benchmark \cite{Wang2018-fr}. Of those reviewed for this paper, only UNIMO and OFA were explicitly benchmarked on such pure NLP tasks.

\section{Pretraining Data}
\label{sec:data}

In this section we address the subjects of the source and size of the various pretraining datasets used to train the models we've discussed. Though not intrinsically related to the models, the source and amount of pretraining data have a tremendous impact on the models downstream performance. The majority of models are pretrained using a small set of publicly available vision language datasets which are briefly described in the subsection immediately below. In the following subsection we provide a similar description of the several models that make use of proprietary datasets. Finally, we end with a subsection devoted to the various sizes of the pretraining datasets used.

\subsection{Public Data Sources}

The following public datasets of the models described above are used alone or in combination with other datasets and form the bulk of the pretraining data for the models that we've discussed. Two of them MSCOCO snd Visual Genome were annotated by human subjects. The remainder were collected and filtered from the web and are in general much larger in scale, but are noisier. 

\subsubsection{Human Annotated Datasets}

\paragraph{MSCOCO} Microsoft's Common Objects in Context (MSCOCO) is among the most widely used datasets for VL tasks. The dataset was introduced by \cite{Lin2014-ju} and was originally intended for object recognition tasks. Overall it consists of a total of 2.5 million labeled object instances in 328k unique images. Of that, over 164,000 of them contain up to 5 independent, user-generated sentences describing the scene. The training split of MSCOCO contains 118k images and their annotations, though most model creators remove some images to avoid data leakage to downstream tasks. ViLT, for instance uses a total of 113k images with 5 captions each for a total of 567k image-text pairs. The annotations were all produced using workers via Amazon's Mechanical Turk service. MSCOCO is publicly available and is used for pretraining, generally in conjunction with other datasets, by a number of the model's we reviewed, including ViLT, METER, mDETR and many others. 

\paragraph{Visual Genome} The Visual Genome dataset \cite{krishna2017visual} consists of only 108k unique images. The images are from the intersection of MSCOCO and YFCC100M \cite{thomee2016yfcc100m}. The set of human derived annotations, also generated using Amazon Mechanical Turk, however is much richer than the MSCOCO dataset. Each image includes a large number of detailed descriptions of various image regions, an extensive set of segmented objects and their labels, as well as a scene graph relating the various objects.  Because each image has 50 descriptions of image regions, it can provide more than 5 million image-text pairs for VL pretraining.  An example of an image from Visual Genome with some examples of bounding boxes, referring expressions and part of the its scene graph is shown if Figure~\ref{fig:vg}. Essentially all of the models that we've reviewed trained with publicly available data make use of Visual Genome as a pretraining dataset. In traditional computer vision modeling, it is also among the most popular datasets for training object detectors.

\begin{figure}[]
\centering
\includegraphics[width=8cm, height=4cm]{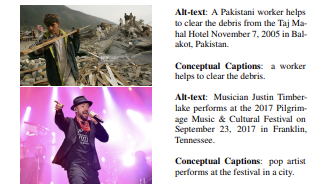}
\caption{From \citet{sharma2018conceptual}}
\label{fig:cc3m}
\end{figure}

\subsubsection{Web Sourced Datasets}

\paragraph{SBU Captions} The Stony Brook University (SBU) Captions dataset was introduced in \citet{ordonez2011im2text} and was one of the first web-scale vision language datasets produced. The images and their associated text were first extracted from Flickr. Then the authors  use a matching process to find additional images without associated text that match those previously collected. Matched photographs are then paired with appropriate text descriptions from the dataset. The process results in a dataset of around 1 million image-text pairs. The SBU captions dataset is used by a number of models in pretraining. It is generally used in conjunction with one or both of the Conceptual Captions sets described below.

\paragraph{Conceptual Captions} The Conceptual Captions dataset was introduced in \citet{sharma2018conceptual}. The data was gathered by crawling a large number of english language websites and filtering for images with accompanying alt-text. The alt-text is then filtered and cleaned, resulting in a concise caption to describe each image. The original Conceptual Captions dataset contains three million image text-pairs and is often referred to as CC3M. Some example images, alt-texts and captions are from the CC3M set are shown in Figure~\ref{fig:cc3m}. The Conceptual 12M dataset was introduced in \citet{changpinyo2021conceptual}. The authors used a similar collection process as \citet{sharma2018conceptual} but were able to dramatically increase the size of the resulting dataset by relaxing the filtering criteria. This increase in scale however, comes at the expense of a much noisier dataset. 

\paragraph{LAION-400M} The LAION-400M dataset \cite{schuhmann2021laion} is a 
 massive collection of images and associated alt-text sourced from the web. The images are filtered for explicit content and caption length. Then a similarity score between each image and its associated text using CLIP; if the score falls below a certain threshold the pair is discarded. In this way, the creators produce a dataset of over 400 million image-text pairs. 

\subsection{Proprietary Datasets} 

\begin{figure}[]
\centering
\includegraphics[width=6cm, height=4cm]{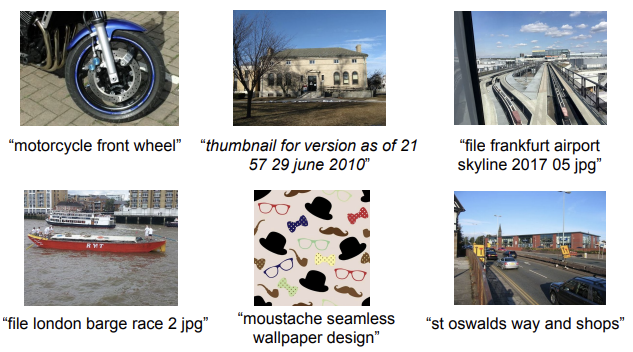}
\caption{From \citet{jia2021scaling}}
\label{fig:align}
\end{figure}

\paragraph{ALIGN} The creators of ALIGN \cite{jia2021scaling} constructed a proprietary dataset consisting of 1.8 billion image-text pairs using the same process that \citet{sharma2018conceptual} used in creating the CC3M dataset. The massive size of the dataset stems from the fact that its creators chose to relax most of the cleaning standards used to create the Conceptual Captions set. Though the resulting data is extremely noisy, the dataset's creators argue that the sheer amount of data tends to overcome the inherent noise. Example image-text pairs from the ALIGN set are shown in Figure~\ref{fig:align}.

\paragraph{CLIP} The CLIP model was trained on a proprietary dataset consisting of 400 million image-text pairs. The model's creator's don't give much detail on their collection process, other than saying "we constructed a new dataset of 400 million (image, text) pairs collected form a variety of publicly available sources on the Internet." 

\paragraph{Flamingo} In addition to the using the dataset from the ALIGN model \cite{jia2021scaling}, the creators of Flamingo also make use of two datasets of their own creations. The first of these is called LTIP (Long Text and Image Pairs) and it consists of 312 million image-text pairs. The authors also curate a custom video-text dataset with 27 million examples. 

\paragraph{LiT} The creators of LiT, created a dataset using the essentially the same process as ALIGN, however, the further relaxed the text-filtering standards. The result is a turly massive dataset consisting of 4 billion image-text pairs.

\subsection{Data Sizes}

In addition to the various sources of pretraining data, we also consider the size of the pretraining dataset used. The size of the pretraining datasets range from that of VisualBERT, which only uses the 567k image-text pairs from COCO, to the 4 billion image-text pairs used to train the LiT model. For convenience we categorize pretraining dataset sizes into small, medium and large sizes. Because we do not currently have the type of extensive meta-analysis studies which formally analyze the effects of pretraining data quantity, these categories are somewhat arbitrary. Said categories are, however, useful in describing distinctions in pretraining data quantities and we begin with models using small datasets, which we define as 10 million or fewer image-text pairs.

\paragraph{Small: Fewer than 10M I-T Pairs} The majority of the models discussed here were trained with 10 million or fewer image-text pairs. The combined number of I-T pairs in the popular datasets MSCOCO, Visual Genome, SBU Captions and CC3M is just under 10 million pairs and a number of models, including BridgeTower, METER, UNIMO, UNITER, VILLA and ViLT, use exactly this dataset. Many models, use even fewer images and only make use of COCO and Visual Genome. E2E-VLP, KD-VLP, PixelBERT use only the 6 million pairs found in MSCOCO and Visual Genome. Because of the need for bounding box annotations in object detection and visual grounding, models that specialize in such tasks make use of the smaller human-annotated datasets designed for said purpose. mDETR, DQ-DETR and KD-VLP for instance use only 1.3 million image-text pairs from Visual Genome, MSCOCO and the Flickr30K Entities set. 

\paragraph{Medium: 10-25M I-T Pairs} By adding the CC12M dataset to corpus of MSCOCO, Visual Genome, SBU Captions and CC3M, we arrive at a dataset with over 21M image-text pairs. This set is used by ALBEF, BEiT-3, mPLUG, OFA and OmniVL for pretraining.

\paragraph{Large: More than 25M I-T Pairs} The dual encoder type models all use relatively large pretraining datasets. CLIP, ALIGN, LiT and Florence all use large web-sourced datasets of 400 million to 4 billion image-text pairs. Several of the large models, such as GIT, PaLI, FLAVA, Flamingo, CoCa and DaVinci use large proprietary datasets. These last models are all quite large and likely need significant amounts of data to leverage the power of their scale.

\section{Analysis and Future Directions}
\label{sec:conclusions}

In this final section, we draw some general conclusions about the models we've previously discussed. The first two sections are devoted to the overall strengths and limitations of the available VL transformer models. In the third section, we discuss some of the open questions and possible directions for future research. Finally, we conclude the paper with a concise statement on the state of VL transformers and their likely role in the immediate future of AI research.  

 

\subsection{Strengths}

\subsubsection{Generalized Representations}

One of the principal advantages of transfer learning from pretrained models is the incredible adaptability they offer. Prior to the recent advent of pretrained VL transformers, complex, task-specific models were required to reach state-of-the-art performance on popular VL benchmarks. Examples include DGAF \cite{gao2019dynamic}, for visual question answering, MAttNet \cite{DBLP:journals/corr/abs-1801-08186}, designed to perform referring expression related tasks, and R2C \cite{Zellers2018-nm}, for visual common sense reasoning. Each model is designed and trained for the end task and could not be easily adapted to other tasks, whereas all of the fusion encoder and encoder-decoder models we have described can be easily transferred to other tasks with minimal changes in architecture and manageable training requirements. Though the pretraining cost of these models is very high (often excessively so), it is incurred only once and the models can be fine-tuned with significantly less compute and limited data input required.

\subsubsection{Ease of Understanding and Use}

In addition to being confined to a small set of use cases, task-specific VL models are often complex and difficult to comprehend. Consider the MAttNet model \cite{DBLP:journals/corr/abs-1801-08186}), designed to perform the reference resolution task (i.e. given an expression describing an object in an image, the model must identify the object referred to). The MAttNet model decomposes word embeddings produced by an RNN model into modules describing different aspects of a phrase and uses an attention network to relate them to image regions produced by an R-CNN. Understanding and reproducing a model like this would require a great deal of background study and programming experience. Though transformers are themselves complex models, their ubiquity ensure that most deep learning practitioners have some familiarity with their workings.

VL-Transformers also simplify the practical use of vision language modeling. After the costly pretraining phase has been completed, most VL-transformers can be adapted to new tasks by a straight forward fine-tuning process. Fine-tuning generally consists of removing the final layer of the model and replacing it with a randomly initialized linear layer with an activation appropriate to the task at hand, such as a softmax or sigmoid layer for classification problems. The weights are then updated as a traditional supervised learning task. The LAVIS \cite{li2022lavis} vision language framework further simplifies the process. LAVIS is Python library for VL modeling that abstracts much of the difficult programming associated with deploying VL models and includes models such as CLIP, BLIP and ALBEF. With efforts such as these, VL transformers offer the possibility of making VL tasks as accessible as NLP and CV tasks currently are using pretrained models from each field.

\subsubsection{Performance}

Finally and likely most importantly,  pretrained VL transformers simply perform better on most tasks than task-specific models. As appealing as their adaptability and simplicity are, pretrained transformers would not be so popular if they were routinely outperformed by other models. Just as it did in NLP, the transformer architecture has spurred huge advances in the performance of VL models. Leader boards for popular VL benchmarks as diverse as VQA and referring expression comprehension with RefCOCO are now dominated by transformer models.\footnote{https://eval.ai/web/challenges/challenge-page/830/leaderboard/2278, https://paperswithcode.com/sota/referring-expression-comprehension-on-refcoco}. 

\subsection{Limitations and Open Questions}

\subsubsection{Pretraining Data and Compute}

The tremendous strengths of pretrained transformers come at the cost of huge pretraining data requirements and enormous compute costs. The excessive compute costs arise from the enormous absolute size of VL-Transformers. At the small end are early encoder only models such as UNITER-Base, which contains 83M parameters. Even these "small" models are quite large and require enormous compute resources to pretrain or fine-tune. This problem is compounded during model development and research since multiple models are usually trained to test various configurations and hyper-parameters and to ensure that results are reproducible. As part of a meta-analysis experiment for VL transformers, \citet{bugliarello2021multimodal} created a controlled setup for analyzing model performance. They estimated that training a single VL transformer 10 times for 4 downstream tasks required a 4-GPU machine on AWS for two months, costing around \$6,000 at the time they published the paper in 2021. On the other end of the spectrum are models based on large language models such as Flamingo which contains 80B parameters. Models this size require enormous compute simply for the purpose of inference, to say nothing of training.

A related problem that transformers face is the amount of pretraining data that they require to achieve strong results. Pretraining data requirements for VL transformers range from that of models like ViLBERT, which uses 3 million image-text pairs from the Conceptual Captions dataset, to GIT-2 which uses a whopping 12.9B imag-text pairs. Simply storing and processing this many image-text pairs is a cumbersome process.  The need for massive datasets is aggravated by the fact that image-text data is difficult to produce.  In order to be useful, the text data must be somehow related to the image it is paired with. Creating datasets of this kind often requires human annotation, though such set are generally quite small given the expense involved. Web-crawled datasets, such as the 1.8B image-text pairs used to train ALIGN, can be obtained using images and their assiciated alt-text, though these data are much noisier than human annotated data.

\subsubsection{Pretraining Tasks}

Independent of the amount of data and compute required to pretrain VL models, it is difficult to find training tasks that explicitly align vision and language and determine how much they contribute to performance. Extending the masked language modeling (or traditional language modeling) task to include visual elements is one strategy pursued by most of the models we have discussed in this paper. Yet standard NLP models such as BERT are already very good at the language modeling without the use of visual information. It is difficult then to determine how much visual feature representations actually contribute to a given task. This is especially true for models such as ViLBERT where the language processing stream is initialized with the pretrained weights from the standard BERT NLP model \cite{devlin2018bert}. 

To explicitly model vision and language interaction, many of the models we have discussed use the image-text matching task we described in Section~\ref{subsec:itm}. Though this task has the advantage of requiring that vision and language interaction be modeled, it does not require the model to make connections between the meaning of words and the objects they actually describe. Furthermore, the image-text matching objective can be viewed as an extension of the next sentence prediction used in the training of BERT. Subsequent work, such as \citet{liu2019roberta}, showed that the next sentence objective can be eliminated without hurting the model's performance. It is worth investigating whether such a straightforward binary classification is also providing marginal contributions to the VL pretraining process. 

A variety of models do make use of more involved pretraining tasks. LXMERT, OFA and VL-T5 for instance all make use of a visual question answering task. Visual grounding tasks, matching objects in images to textual descriptions, are used by OFA, VL-T5, mDETR, X2-VLM and several other models. Pretraining tasks of this nature more explicitly align vision and language and could possibly prove superior as proxy tasks. Unfortunately, tasks of this nature often require specialized datasets. Visual grounding tasks, for instance require the type of annotations found in RefCOCO. Said annotations are expensive to produce and the resulting dataset is very small. This poses an acute problem for pretraining data hungry transformers. Furthermore, to date, there is no detailed model development process or meta-analysis showing to what degree  pretraining regimes such as these affect downstream performance. Until such data is available in the literature, it will remain an open question what a maximally effective vision language pretraining regime would entail.


\subsubsection{Visual Embeddings}

It is still an open question, as to what visual embedding strategy is best and under what circumstances. Many early models, ViLBERT and UNITER among them, made use of the region features described in Section \ref{sec:embedding}. These features were a popular choice for visual embedding type because they can be produced by out of the box R-CNN models and directly fed into VL models, yet they have several drawbacks. Firstly, region features are confined to the object categories on which the object detection model was trained. This places an inherent limit on what types of visual information that the model can learn. Secondly, region features are usually denoted by rectangular bounding boxes and don't account well for shapes of objects they contain or their relationship to one another. Finally, CNN-based object detection models consume a great deal of compute and represent a serious bottleneck in the modeling process. 

The creators of ViLT \cite{kim2021vilt} performed an analysis of the inference time for several of the models we discussed. Using region features, fusion encoders like the UNITER model spend the overwhelming majority of their inference time creating and operating on visual embeddings. Beyond creating a computational bottleneck, spending more time on data preprocessing than modeling is simply not a desirable design feature. 

The grid features of PixelBERT \cite{huang2020pixel} and SOHO \cite{Huang2021-ee} have the advantages of reducing visual processing requirements and removing the theoretical ceiling on visual learning of region features. They also offer a dense set of image features for each image input. But, a model like PixelBERT still spends more time on visual preprocessing than VL modeling \cite{kim2021vilt}. Furthermore, a separate CNN module was required to produce 
grid features, adding additional complexity to the model. 

Finally, models such as ViLT, that use patch embeddings spend a negligible percentage of inference time on visual processing. While this appears to solve the issue of excessive compute requirements for processing visual features, there is some concern that patch emeddings provide inferior representatoins to other approaches. ViLT, the first model found in the literature to make use of patch embeddings performed worse on most VL understanding tasks than similar models using region features. Later models that use patch embeddings, such as METER and BridgeTower, however, perform quite well on these tasks. A potential deficit of visual information is of special concern for object identification tasks where patch embeddings don't have access to the explicit object type information included in region feature representations. Almost all of the models that made use of visual grounding tasks in pretraining used, e.g. mDETR, use grid or region features. Without more detailed meta-analysis of VL transformers, it is still difficult to determine the best approach to visual embeddings.   






\subsection{Future Directions}

\subsubsection{Data Generation and Meta Analysis}

Though not directly tied to the process of model development, the generation of more quality vision language datasets is a clear need in the current research environment. There is currently a deficit of publicly available pretraining data as well as quality datasets for benchmarking VL models. The huge majority of models reviewed for this paper use some combination of MSCOCO, Visual Genome, SBU Captions and one or both of the Conceptual Captions datasets for pretraining. The addition of more quality pretraining data has been very beneficial to transformer's in other domains \cite{liu2019roberta} and would almost certainly improve the performance of most of the models we've described. 

Similarly we also see a paucity of data on which to benchmark model performance. The models we've discussed here all evaluated on a narrow set of downstream tasks such as VQA \cite{Antol2015-cs}, NLVR2 \cite{suhr2018corpus} and the COCO cross modal retrieval task. These VL datasets must have images paired with textual components that accurately describe, question or reference them and are resource intensive to create, usually requiring human annotators. That being said, crowd sourcing services such as Amazon Mechanical Turk greatly simplify the process of rote tasks such as annotating images. Indeed, several important datasets in the VL domain, such as MSCOCO \cite{Lin2014-ju} and VCR \cite{Zellers2018-nm} were created using its services. The process is then reduced to finding images, creating a script for anonymous workers to follow for annotation and compensating workers. Though the need for human annotation makes the cost of creating such datasets relatively high, we have already noted that creating and testing the models discussed here were resource intensive tasks. Well-resourced institutions interested in advancing the field would surely benefit as much from more quality data to evaluate models as from creating additional variations of VL transformers. 

Another obvious priority is understanding the performance of the various models that have already been created. One of the few published papers attempting a meta-analysis of current VL transformer models is \citet{bugliarello2021multimodal}. This analysis sheds some light on how some of the models that we've discussed perform, yet it leaves open important questions such as whether dual or single stream fusion encoders offer superior performance, whether or not the patch embedding strategy is a viable solution to the problems posed by region features, and how effective are the pretraining objectives currently in use. Answering questions of this nature would require thorough controlled studies in which potential confounds such as pretraining data size (and type), training hyper-parameters that are held constant, and each model configuration is trained multiple times and evaluated on a broad range of downstream tasks. Studies of this nature would be a significant undertaking with such large models. 

As with data generation, however, investments in meta-analysis should surely take priority over model development. Simply put, model development has outpaced the supply of data and what we know about how these models perform. If dual-tower architectures cannot be shown to improve performance, it is hard to justify the additional parameters of a second transformer encoder stack. Similarly, if patch embeddings can perform on par with region features, they represent a better visual embedding strategy based on their huge advantage in efficiency. Along with controlled experimentation, a large and diverse body of data for model pretraining and evaluation are the keys to resolving these open questions and guiding future research.

\subsubsection{Alternative Pretraining Tasks}

The pretraining regimes that we've covered in this paper are mostly straightforward extensions of the pretraining objectives used in NLP models such as BERT. The choice to extend BERT's training objectives was a natural place to start, however, as we have previously described, there seems to be a great deal of room for exploring new pretraining objectives that create deeper and more explicit interactions between vision and language modalities. In particular, we believe that VL transformers would benefit from a more complex and detailed set of training objectives such as visual grounding tasks and visual question answering. 

In the course of developing the METER model, \citet{dou2022empirical} conducted an extensive study of various model architectures in search of the most effective design. We propose that a similar approach testing novel pretraining tasks would also be beneficial and could possibly result in state-of-the-art results. In addition to systematically testing the pretraining objectives that we have previously introduced, there are other approaches worth investigating. Position guided text prompting \cite{wang2023position} is an example of such a task. It splits an image into patches, identifies various objects in each patch and produces a series of fill in the blank tasks based on said object information.  Crucially, the authors designed it to be compatible (at least theoretically so) with any type of VL transformer architecture. A thorough investigation of such pretraining tasks would almost surely move the field of VL modeling forward.  

\subsubsection{Additional Modalities}

The extraordinary  versatility of transformers recommends them to be applied to additional modalities beyond just language and vision. An evolving body of research is attempting to do just that. Models such as ONE-PEACE \cite{wang2023one} and VALOR \cite{chen2023valor} model vision and language but also incorporate the audio modality using techniques that are quite similar to the ones discussed in this paper.  Remarkably, transformers can also be applied to even more sensory modalities. PALM-E \cite{driess2023palm} is an embodied vision language models that is capable of performing a variety of robot maniputlation tasks. The model can additionally perform the types of vision language tasks referenced throughout this paper as well as language-only tasks. In addition to the obvious practical applications that these more general models might have, they are offer a possible path toward resolving problems like the symbol grounding problem described in \citet{Harnad1990-ws}.  Language models that incorporate both audio and video modalities are one step closer to the type of language grounding that underscores human speech. 

\subsection{Concluding Remarks}

Pretrained VL transformer models are a relatively new creations and there is still a great deal of research to be done to determine how they should be designed and trained. It is not yet clear which of the several approaches we've discussed are superior. This includes basic questions such as how to create visual embeddings, single stream versus dual-tower architectures or which pretraining objectives to use and when. Pretrained transformers also come at the cost of huge pretraining data requirements, and this is of particular concern in the VL domain, where suitable data is hard to find and expensive to produce. Pretraining these models would probably also benefit from a new set of training tasks that explicitly align language and vision in deep and meaningful ways. 

With all of that said, the models discussed here do appear to considerably advance the state-of-the-art on the most common VL benchmarks currently available.  Moreover, they do this via generalized models that can be easily adapted to a variety of tasks.  VL transformers also offer the possibility of being extended to other sensory modalities. In general, they represent a remarkable step forward from previous VL models. Connecting symbols such as words to their real-world analogs is at the core of intelligence \cite{Harnad1990-ws} and the models we've discussed offer promising paths toward achieving this goal and significantly improving NLP and AI in general. Given these remarkable strengths, pretrained VL transformers will likely play a key role in the near future of modeling tasks where vision and language intersect.




\bibliography{custom,paperpile}

\begin{thebibliography}{95}
\expandafter\ifx\csname natexlab\endcsname\relax\def\natexlab#1{#1}\fi

\bibitem[{Agrawal et~al.(2019)Agrawal, Desai, Wang, Chen, Jain, Johnson, Batra,
  Parikh, Lee, and Anderson}]{Agrawal2019-lv}
Harsh Agrawal, Karan Desai, Yufei Wang, Xinlei Chen, Rishabh Jain, Mark
  Johnson, Dhruv Batra, Devi Parikh, Stefan Lee, and Peter Anderson. 2019.
\newblock Nocaps: Novel object captioning at scale.
\newblock In \emph{2019 {IEEE/CVF} International Conference on Computer Vision
  ({ICCV})}. IEEE.

\bibitem[{Alayrac et~al.(2022)Alayrac, Donahue, Luc, Miech, Barr, Hasson, Lenc,
  Mensch, Millican, Reynolds, Ring, Rutherford, Cabi, Han, Gong, Samangooei,
  Monteiro, Menick, Borgeaud, Brock, Nematzadeh, Sharifzadeh, Binkowski,
  Barreira, Vinyals, Zisserman, and Simonyan}]{Alayrac2022-yp}
Jean-Baptiste Alayrac, Jeff Donahue, Pauline Luc, Antoine Miech, Iain Barr,
  Yana Hasson, Karel Lenc, Arthur Mensch, Katie Millican, Malcolm Reynolds,
  Roman Ring, Eliza Rutherford, Serkan Cabi, Tengda Han, Zhitao Gong, Sina
  Samangooei, Marianne Monteiro, Jacob Menick, Sebastian Borgeaud, Andrew
  Brock, Aida Nematzadeh, Sahand Sharifzadeh, Mikolaj Binkowski, Ricardo
  Barreira, Oriol Vinyals, Andrew Zisserman, and Karen Simonyan. 2022.
\newblock \href {http://arxiv.org/abs/2204.14198} {Flamingo: A visual language
  model for few-shot learning}.

\bibitem[{Antol et~al.(2015)Antol, Agrawal, Lu, Mitchell, Batra, Zitnick, and
  Parikh}]{Antol2015-cs}
Stanislaw Antol, Aishwarya Agrawal, Jiasen Lu, Margaret Mitchell, Dhruv Batra,
  C~Lawrence Zitnick, and Devi Parikh. 2015.
\newblock {VQA}: Visual question answering.
\newblock In \emph{2015 {IEEE} International Conference on Computer Vision
  ({ICCV})}. IEEE.

\bibitem[{Bao et~al.(2021)Bao, Dong, and Wei}]{bao2021beit}
Hangbo Bao, Li~Dong, and Furu Wei. 2021.
\newblock Beit: Bert pre-training of image transformers.
\newblock \emph{arXiv preprint arXiv:2106.08254}.

\bibitem[{Bao et~al.(2022)Bao, Wang, Dong, Liu, Mohammed, Aggarwal, Som, Piao,
  and Wei}]{bao2022vlmo}
Hangbo Bao, Wenhui Wang, Li~Dong, Qiang Liu, Owais~Khan Mohammed, Kriti
  Aggarwal, Subhojit Som, Songhao Piao, and Furu Wei. 2022.
\newblock Vlmo: Unified vision-language pre-training with
  mixture-of-modality-experts.
\newblock \emph{Advances in Neural Information Processing Systems},
  35:32897--32912.

\bibitem[{Brown et~al.(2020)Brown, Mann, Ryder, Subbiah, Kaplan, Dhariwal,
  Neelakantan, Shyam, Sastry, Askell et~al.}]{brown2020language}
Tom Brown, Benjamin Mann, Nick Ryder, Melanie Subbiah, Jared~D Kaplan, Prafulla
  Dhariwal, Arvind Neelakantan, Pranav Shyam, Girish Sastry, Amanda Askell,
  et~al. 2020.
\newblock Language models are few-shot learners.
\newblock \emph{Advances in neural information processing systems},
  33:1877--1901.

\bibitem[{Bugliarello et~al.(2021)Bugliarello, Cotterell, Okazaki, and
  Elliott}]{bugliarello2021multimodal}
Emanuele Bugliarello, Ryan Cotterell, Naoaki Okazaki, and Desmond Elliott.
  2021.
\newblock Multimodal pretraining unmasked: A meta-analysis and a unified
  framework of vision-and-language berts.
\newblock \emph{Transactions of the Association for Computational Linguistics},
  9:978--994.

\bibitem[{Carion et~al.(2020)Carion, Massa, Synnaeve, Usunier, Kirillov, and
  Zagoruyko}]{Carion2020-hb}
Nicolas Carion, Francisco Massa, Gabriel Synnaeve, Nicolas Usunier, Alexander
  Kirillov, and Sergey Zagoruyko. 2020.
\newblock End-to-end object detection with transformers.
\newblock In \emph{Computer Vision -- {ECCV} 2020}, Lecture notes in computer
  science, pages 213--229. Springer International Publishing, Cham.

\bibitem[{Changpinyo et~al.(2021)Changpinyo, Sharma, Ding, and
  Soricut}]{changpinyo2021conceptual}
Soravit Changpinyo, Piyush Sharma, Nan Ding, and Radu Soricut. 2021.
\newblock Conceptual 12m: Pushing web-scale image-text pre-training to
  recognize long-tail visual concepts.
\newblock In \emph{Proceedings of the IEEE/CVF Conference on Computer Vision
  and Pattern Recognition}, pages 3558--3568.

\bibitem[{Chen et~al.(2023)Chen, He, Guo, Zhu, Wang, Tang, and
  Liu}]{chen2023valor}
Sihan Chen, Xingjian He, Longteng Guo, Xinxin Zhu, Weining Wang, Jinhui Tang,
  and Jing Liu. 2023.
\newblock Valor: Vision-audio-language omni-perception pretraining model and
  dataset.
\newblock \emph{arXiv preprint arXiv:2304.08345}.

\bibitem[{Chen et~al.(2022)Chen, Wang, Changpinyo, Piergiovanni, Padlewski,
  Salz, Goodman, Grycner, Mustafa, Beyer et~al.}]{chen2022pali}
Xi~Chen, Xiao Wang, Soravit Changpinyo, AJ~Piergiovanni, Piotr Padlewski,
  Daniel Salz, Sebastian Goodman, Adam Grycner, Basil Mustafa, Lucas Beyer,
  et~al. 2022.
\newblock Pali: A jointly-scaled multilingual language-image model.
\newblock \emph{arXiv preprint arXiv:2209.06794}.

\bibitem[{Chen et~al.(2015)Chen, Fang, Lin, Vedantam, Gupta, Dollar, and
  Lawrence~Zitnick}]{Chen2015-my}
Xinlei Chen, Hao Fang, Tsung-Yi Lin, Ramakrishna Vedantam, Saurabh Gupta, Piotr
  Dollar, and C~Lawrence~Zitnick. 2015.
\newblock \href {http://arxiv.org/abs/1504.00325} {Microsoft {COCO} captions:
  Data collection and evaluation server}.

\bibitem[{Chen et~al.(2019)Chen, Li, Yu, El~Kholy, Ahmed, Gan, Cheng, and
  Liu}]{Chen2019-sh}
Yen-Chun Chen, Linjie Li, Licheng Yu, Ahmed El~Kholy, Faisal Ahmed, Zhe Gan,
  Yu~Cheng, and Jingjing Liu. 2019.
\newblock {UNITER}: Learning {UNiversal} {Image-TExt} representations.

\bibitem[{Cho et~al.(2021)Cho, Lei, Tan, and Bansal}]{cho2021unifying}
Jaemin Cho, Jie Lei, Hao Tan, and Mohit Bansal. 2021.
\newblock Unifying vision-and-language tasks via text generation.
\newblock In \emph{International Conference on Machine Learning}, pages
  1931--1942. PMLR.

\bibitem[{Dai et~al.(2021)Dai, Liu, Le, and Tan}]{Dai2021-ps}
Zihang Dai, Hanxiao Liu, Quoc~V Le, and Mingxing Tan. 2021.
\newblock \href {http://arxiv.org/abs/2106.04803} {{CoAtNet}: Marrying
  convolution and attention for all data sizes}.

\bibitem[{Devlin et~al.(2018)Devlin, Chang, Lee, and
  Toutanova}]{devlin2018bert}
Jacob Devlin, Ming-Wei Chang, Kenton Lee, and Kristina Toutanova. 2018.
\newblock Bert: Pre-training of deep bidirectional transformers for language
  understanding.
\newblock \emph{arXiv preprint arXiv:1810.04805}.

\bibitem[{Diao et~al.(2022)Diao, Zhou, Zhang, and Wang}]{diao2022prefix}
Shizhe Diao, Wangchunshu Zhou, Xinsong Zhang, and Jiawei Wang. 2022.
\newblock Prefix language models are unified modal learners.
\newblock \emph{arXiv preprint arXiv:2206.07699}.

\bibitem[{Dong et~al.(2021)Dong, Bao, Chen, Zhang, Yu, Yuan, Chen, and
  Guo}]{Dong2021-rj}
Xiaoyi Dong, Jianmin Bao, Dongdong Chen, Weiming Zhang, Nenghai Yu, Lu~Yuan,
  Dong Chen, and Baining Guo. 2021.
\newblock \href {http://arxiv.org/abs/2107.00652} {{CSWin} transformer: A
  general vision transformer backbone with cross-shaped windows}.

\bibitem[{Dosovitskiy et~al.(2020)Dosovitskiy, Beyer, Kolesnikov, Weissenborn,
  Zhai, Unterthiner, Dehghani, Minderer, Heigold, Gelly
  et~al.}]{dosovitskiy2020image}
Alexey Dosovitskiy, Lucas Beyer, Alexander Kolesnikov, Dirk Weissenborn,
  Xiaohua Zhai, Thomas Unterthiner, Mostafa Dehghani, Matthias Minderer, Georg
  Heigold, Sylvain Gelly, et~al. 2020.
\newblock An image is worth 16x16 words: Transformers for image recognition at
  scale.
\newblock \emph{arXiv preprint arXiv:2010.11929}.

\bibitem[{Dou et~al.(2022)Dou, Xu, Gan, Wang, Wang, Wang, Zhu, Zhang, Yuan,
  Peng et~al.}]{dou2022empirical}
Zi-Yi Dou, Yichong Xu, Zhe Gan, Jianfeng Wang, Shuohang Wang, Lijuan Wang,
  Chenguang Zhu, Pengchuan Zhang, Lu~Yuan, Nanyun Peng, et~al. 2022.
\newblock An empirical study of training end-to-end vision-and-language
  transformers.
\newblock In \emph{Proceedings of the IEEE/CVF Conference on Computer Vision
  and Pattern Recognition}, pages 18166--18176.

\bibitem[{Driess et~al.(2023)Driess, Xia, Sajjadi, Lync, Chowdhery, Ichter,
  Wahid, Tompson, Vuong, Yu et~al.}]{driess2023palm}
Danny Driess, Fei Xia, Mehdi~SM Sajjadi, Corey Lync, h, Aakanksha Chowdhery,
  Brian Ichter, Ayzaan Wahid, Jonathan Tompson, Quan Vuong, Tianhe Yu, et~al.
  2023.
\newblock Palm-e: An embodied multimodal language model.
\newblock \emph{arXiv preprint arXiv:2303.03378}.

\bibitem[{Gan et~al.(2020)Gan, Chen, Li, Zhu, Cheng, and Liu}]{gan2020large}
Zhe Gan, Yen-Chun Chen, Linjie Li, Chen Zhu, Yu~Cheng, and Jingjing Liu. 2020.
\newblock Large-scale adversarial training for vision-and-language
  representation learning.
\newblock \emph{Advances in Neural Information Processing Systems},
  33:6616--6628.

\bibitem[{Gao et~al.(2019)Gao, Jiang, You, Lu, Hoi, Wang, and
  Li}]{gao2019dynamic}
Peng Gao, Zhengkai Jiang, Haoxuan You, Pan Lu, Steven~CH Hoi, Xiaogang Wang,
  and Hongsheng Li. 2019.
\newblock Dynamic fusion with intra-and inter-modality attention flow for
  visual question answering.
\newblock In \emph{Proceedings of the IEEE/CVF conference on computer vision
  and pattern recognition}, pages 6639--6648.

\bibitem[{Girshick(2015)}]{girshick2015fast}
Ross Girshick. 2015.
\newblock Fast r-cnn.
\newblock In \emph{Proceedings of the IEEE international conference on computer
  vision}, pages 1440--1448.

\bibitem[{Gupta et~al.(2021)Gupta, Kamath, Kembhavi, and Hoiem}]{Gupta2021-kn}
Tanmay Gupta, Amita Kamath, Aniruddha Kembhavi, and Derek Hoiem. 2021.
\newblock \href {http://arxiv.org/abs/2104.00743} {Towards general purpose
  vision systems}.

\bibitem[{Harnad(1990)}]{Harnad1990-ws}
Stevan Harnad. 1990.
\newblock The symbol grounding problem.
\newblock \emph{Physica D}, 42(1):335--346.

\bibitem[{He et~al.(2015)He, Zhang, Ren, and Sun}]{He2015-et}
Kaiming He, Xiangyu Zhang, Shaoqing Ren, and Jian Sun. 2015.
\newblock \href {http://arxiv.org/abs/1512.03385} {Deep residual learning for
  image recognition}.

\bibitem[{Hu et~al.(2022)Hu, Gan, Wang, Yang, Liu, Lu, and Wang}]{Hu2022-kq}
Xiaowei Hu, Zhe Gan, Jianfeng Wang, Zhengyuan Yang, Zicheng Liu, Yumao Lu, and
  Lijuan Wang. 2022.
\newblock Scaling up vision-language pretraining for image captioning.
\newblock In \emph{2022 {IEEE/CVF} Conference on Computer Vision and Pattern
  Recognition ({CVPR})}. IEEE.

\bibitem[{Huang et~al.(2021{\natexlab{a}})Huang, Zeng, Huang, Liu, Fu, and
  Fu}]{Huang2021-ee}
Zhicheng Huang, Zhaoyang Zeng, Yupan Huang, Bei Liu, Dongmei Fu, and Jianlong
  Fu. 2021{\natexlab{a}}.
\newblock \href {http://arxiv.org/abs/2104.03135} {Seeing out of the box:
  {End-to-End} pre-training for {Vision-Language} representation learning}.

\bibitem[{Huang et~al.(2021{\natexlab{b}})Huang, Zeng, Huang, Liu, Fu, and
  Fu}]{huang2021seeing}
Zhicheng Huang, Zhaoyang Zeng, Yupan Huang, Bei Liu, Dongmei Fu, and Jianlong
  Fu. 2021{\natexlab{b}}.
\newblock Seeing out of the box: End-to-end pre-training for vision-language
  representation learning.
\newblock In \emph{Proceedings of the IEEE/CVF Conference on Computer Vision
  and Pattern Recognition}, pages 12976--12985.

\bibitem[{Huang et~al.(2020)Huang, Zeng, Liu, Fu, and Fu}]{huang2020pixel}
Zhicheng Huang, Zhaoyang Zeng, Bei Liu, Dongmei Fu, and Jianlong Fu. 2020.
\newblock Pixel-bert: Aligning image pixels with text by deep multi-modal
  transformers.
\newblock \emph{arXiv preprint arXiv:2004.00849}.

\bibitem[{Jia et~al.(2021)Jia, Yang, Xia, Chen, Parekh, Pham, Le, Sung, Li, and
  Duerig}]{jia2021scaling}
Chao Jia, Yinfei Yang, Ye~Xia, Yi-Ting Chen, Zarana Parekh, Hieu Pham, Quoc Le,
  Yun-Hsuan Sung, Zhen Li, and Tom Duerig. 2021.
\newblock Scaling up visual and vision-language representation learning with
  noisy text supervision.
\newblock In \emph{International Conference on Machine Learning}, pages
  4904--4916. PMLR.

\bibitem[{Joshi et~al.(2020)Joshi, Chen, Liu, Weld, Zettlemoyer, and
  Levy}]{joshi2020spanbert}
Mandar Joshi, Danqi Chen, Yinhan Liu, Daniel~S Weld, Luke Zettlemoyer, and Omer
  Levy. 2020.
\newblock Spanbert: Improving pre-training by representing and predicting
  spans.
\newblock \emph{Transactions of the Association for Computational Linguistics},
  8:64--77.

\bibitem[{Kamath et~al.(2021)Kamath, Singh, LeCun, Synnaeve, Misra, and
  Carion}]{kamath2021mdetr}
Aishwarya Kamath, Mannat Singh, Yann LeCun, Gabriel Synnaeve, Ishan Misra, and
  Nicolas Carion. 2021.
\newblock Mdetr-modulated detection for end-to-end multi-modal understanding.
\newblock In \emph{Proceedings of the IEEE/CVF International Conference on
  Computer Vision}, pages 1780--1790.

\bibitem[{Kazemzadeh et~al.(2014)Kazemzadeh, Ordonez, Matten, and
  Berg}]{kazemzadeh2014referitgame}
Sahar Kazemzadeh, Vicente Ordonez, Mark Matten, and Tamara Berg. 2014.
\newblock Referitgame: Referring to objects in photographs of natural scenes.
\newblock In \emph{Proceedings of the 2014 conference on empirical methods in
  natural language processing (EMNLP)}, pages 787--798.

\bibitem[{Kim et~al.(2021)Kim, Son, and Kim}]{kim2021vilt}
Wonjae Kim, Bokyung Son, and Ildoo Kim. 2021.
\newblock Vilt: Vision-and-language transformer without convolution or region
  supervision.
\newblock In \emph{International Conference on Machine Learning}, pages
  5583--5594. PMLR.

\bibitem[{Krishna et~al.(2017)Krishna, Zhu, Groth, Johnson, Hata, Kravitz,
  Chen, Kalantidis, Li, Shamma et~al.}]{krishna2017visual}
Ranjay Krishna, Yuke Zhu, Oliver Groth, Justin Johnson, Kenji Hata, Joshua
  Kravitz, Stephanie Chen, Yannis Kalantidis, Li-Jia Li, David~A Shamma, et~al.
  2017.
\newblock Visual genome: Connecting language and vision using crowdsourced
  dense image annotations.
\newblock \emph{International journal of computer vision}, 123(1):32--73.

\bibitem[{Kudo and Richardson(2018)}]{kudo2018sentencepiece}
Taku Kudo and John Richardson. 2018.
\newblock Sentencepiece: A simple and language independent subword tokenizer
  and detokenizer for neural text processing.
\newblock \emph{arXiv preprint arXiv:1808.06226}.

\bibitem[{Lan et~al.(2019)Lan, Chen, Goodman, Gimpel, Sharma, and
  Soricut}]{lan2019albert}
Zhenzhong Lan, Mingda Chen, Sebastian Goodman, Kevin Gimpel, Piyush Sharma, and
  Radu Soricut. 2019.
\newblock Albert: A lite bert for self-supervised learning of language
  representations.
\newblock \emph{arXiv preprint arXiv:1909.11942}.

\bibitem[{Li et~al.(2022{\natexlab{a}})Li, Xu, Tian, Wang, Yan, Bi, Ye, Chen,
  Xu, Cao et~al.}]{li2022mplug}
Chenliang Li, Haiyang Xu, Junfeng Tian, Wei Wang, Ming Yan, Bin Bi, Jiabo Ye,
  Hehong Chen, Guohai Xu, Zheng Cao, et~al. 2022{\natexlab{a}}.
\newblock mplug: Effective and efficient vision-language learning by
  cross-modal skip-connections.
\newblock \emph{arXiv preprint arXiv:2205.12005}.

\bibitem[{Li et~al.(2022{\natexlab{b}})Li, Li, Le, Wang, Savarese, and
  Hoi}]{li2022lavis}
Dongxu Li, Junnan Li, Hung Le, Guangsen Wang, Silvio Savarese, and Steven C.~H.
  Hoi. 2022{\natexlab{b}}.
\newblock \href {http://arxiv.org/abs/2209.09019} {Lavis: A library for
  language-vision intelligence}.

\bibitem[{Li et~al.(2020{\natexlab{a}})Li, Duan, Fang, Gong, and
  Jiang}]{Li2020-ea}
Gen Li, Nan Duan, Yuejian Fang, Ming Gong, and Daxin Jiang. 2020{\natexlab{a}}.
\newblock {Unicoder-VL}: A universal encoder for vision and language by
  {Cross-Modal} {Pre-Training}.
\newblock \emph{AAAI}, 34(07):11336--11344.

\bibitem[{Li et~al.(2023)Li, Li, Savarese, and Hoi}]{li2023blip}
Junnan Li, Dongxu Li, Silvio Savarese, and Steven Hoi. 2023.
\newblock Blip-2: Bootstrapping language-image pre-training with frozen image
  encoders and large language models.
\newblock \emph{arXiv preprint arXiv:2301.12597}.

\bibitem[{Li et~al.(2021)Li, Selvaraju, Gotmare, Joty, Xiong, and
  Hoi}]{li2021align}
Junnan Li, Ramprasaath Selvaraju, Akhilesh Gotmare, Shafiq Joty, Caiming Xiong,
  and Steven Chu~Hong Hoi. 2021.
\newblock Align before fuse: Vision and language representation learning with
  momentum distillation.
\newblock \emph{Advances in neural information processing systems},
  34:9694--9705.

\bibitem[{Li et~al.(2019)Li, Yatskar, Yin, Hsieh, and Chang}]{li2019visualbert}
Liunian~Harold Li, Mark Yatskar, Da~Yin, Cho-Jui Hsieh, and Kai-Wei Chang.
  2019.
\newblock Visualbert: A simple and performant baseline for vision and language.
\newblock \emph{arXiv preprint arXiv:1908.03557}.

\bibitem[{Li and Sigal(2021)}]{Li2021-wn}
Muchen Li and Leonid Sigal. 2021.
\newblock \href {http://arxiv.org/abs/2106.03089} {Referring transformer: A
  one-step approach to multi-task visual grounding}.

\bibitem[{Li et~al.(2020{\natexlab{b}})Li, Gao, Niu, Xiao, Liu, Liu, Wu, and
  Wang}]{Li2020-up}
Wei Li, Can Gao, Guocheng Niu, Xinyan Xiao, Hao Liu, Jiachen Liu, Hua Wu, and
  Haifeng Wang. 2020{\natexlab{b}}.
\newblock \href {http://arxiv.org/abs/2012.15409} {{UNIMO}: Towards
  {Unified-Modal} understanding and generation via {Cross-Modal} contrastive
  learning}.

\bibitem[{Li et~al.(2020{\natexlab{c}})Li, Yin, Li, Zhang, Hu, Zhang, Wang, Hu,
  Dong, Wei et~al.}]{li2020oscar}
Xiujun Li, Xi~Yin, Chunyuan Li, Pengchuan Zhang, Xiaowei Hu, Lei Zhang, Lijuan
  Wang, Houdong Hu, Li~Dong, Furu Wei, et~al. 2020{\natexlab{c}}.
\newblock Oscar: Object-semantics aligned pre-training for vision-language
  tasks.
\newblock In \emph{Computer Vision--ECCV 2020: 16th European Conference,
  Glasgow, UK, August 23--28, 2020, Proceedings, Part XXX 16}, pages 121--137.
  Springer.

\bibitem[{Lin et~al.(2014)Lin, Maire, Belongie, Hays, Perona, Ramanan,
  Doll{\'a}r, and Zitnick}]{Lin2014-ju}
Tsung-Yi Lin, Michael Maire, Serge Belongie, James Hays, Pietro Perona, Deva
  Ramanan, Piotr Doll{\'a}r, and C~Lawrence Zitnick. 2014.
\newblock Microsoft {COCO}: Common objects in context.
\newblock In \emph{Computer Vision -- {ECCV} 2014}, pages 740--755. Springer
  International Publishing.

\bibitem[{Liu et~al.(2022{\natexlab{a}})Liu, Liang, Li, Huang, Zhang, Su, Zhu,
  and Zhang}]{Liu2022-pg}
Shilong Liu, Yaoyuan Liang, Feng Li, Shijia Huang, Hao Zhang, Hang Su, Jun Zhu,
  and Lei Zhang. 2022{\natexlab{a}}.
\newblock \href {http://arxiv.org/abs/2211.15516} {{DQ-DETR}: Dual query
  detection transformer for phrase extraction and grounding}.

\bibitem[{Liu et~al.(2019)Liu, Ott, Goyal, Du, Joshi, Chen, Levy, Lewis,
  Zettlemoyer, and Stoyanov}]{liu2019roberta}
Yinhan Liu, Myle Ott, Naman Goyal, Jingfei Du, Mandar Joshi, Danqi Chen, Omer
  Levy, Mike Lewis, Luke Zettlemoyer, and Veselin Stoyanov. 2019.
\newblock Roberta: A robustly optimized bert pretraining approach.
\newblock \emph{arXiv preprint arXiv:1907.11692}.

\bibitem[{Liu et~al.(2022{\natexlab{b}})Liu, Wu, Tseng, Lal, He, and
  Duan}]{Liu2022-qj}
Yongfei Liu, Chenfei Wu, Shao-Yen Tseng, Vasudev Lal, Xuming He, and Nan Duan.
  2022{\natexlab{b}}.
\newblock {KD-VLP}: Improving end-to-end vision-and-language pretraining with
  object knowledge distillation.
\newblock In \emph{Findings of the Association for Computational Linguistics:
  {NAACL} 2022}, Stroudsburg, PA, USA. Association for Computational
  Linguistics.

\bibitem[{Liu et~al.(2021)Liu, Lin, Cao, Hu, Wei, Zhang, Lin, and
  Guo}]{Liu2021-vy}
Ze~Liu, Yutong Lin, Yue Cao, Han Hu, Yixuan Wei, Zheng Zhang, Stephen Lin, and
  Baining Guo. 2021.
\newblock Swin transformer: Hierarchical vision transformer using shifted
  windows.
\newblock In \emph{2021 {IEEE/CVF} International Conference on Computer Vision
  ({ICCV})}. IEEE.

\bibitem[{{Lu} et~al.(2019){Lu}, {Batra}, {Parikh}, and {Lee}}]{Lu2019-bb}
{Lu}, {Batra}, {Parikh}, and {Lee}. 2019.
\newblock Vilbert: Pretraining task-agnostic visiolinguistic representations
  for vision-and-language tasks.
\newblock \emph{Adv. Neural Inf. Process. Syst.}

\bibitem[{OpenAI(2023)}]{openai2023gpt4}
OpenAI. 2023.
\newblock \href {http://arxiv.org/abs/2303.08774} {Gpt-4 technical report}.

\bibitem[{Ordonez et~al.(2011)Ordonez, Kulkarni, and Berg}]{ordonez2011im2text}
Vicente Ordonez, Girish Kulkarni, and Tamara Berg. 2011.
\newblock Im2text: Describing images using 1 million captioned photographs.
\newblock \emph{Advances in neural information processing systems}, 24.

\bibitem[{Peng et~al.(2022)Peng, Dong, Bao, Ye, and Wei}]{peng2022beit}
Zhiliang Peng, Li~Dong, Hangbo Bao, Qixiang Ye, and Furu Wei. 2022.
\newblock Beit v2: Masked image modeling with vector-quantized visual
  tokenizers.
\newblock \emph{arXiv preprint arXiv:2208.06366}.

\bibitem[{Plummer et~al.(2015)Plummer, Wang, Cervantes, Caicedo, Hockenmaier,
  and Lazebnik}]{plummer2015flickr30k}
Bryan~A Plummer, Liwei Wang, Chris~M Cervantes, Juan~C Caicedo, Julia
  Hockenmaier, and Svetlana Lazebnik. 2015.
\newblock Flickr30k entities: Collecting region-to-phrase correspondences for
  richer image-to-sentence models.
\newblock In \emph{Proceedings of the IEEE international conference on computer
  vision}, pages 2641--2649.

\bibitem[{Radford et~al.(2021)Radford, Kim, Hallacy, Ramesh, Goh, Agarwal,
  Sastry, Askell, Mishkin, Clark et~al.}]{radford2021learning}
Alec Radford, Jong~Wook Kim, Chris Hallacy, Aditya Ramesh, Gabriel Goh,
  Sandhini Agarwal, Girish Sastry, Amanda Askell, Pamela Mishkin, Jack Clark,
  et~al. 2021.
\newblock Learning transferable visual models from natural language
  supervision.
\newblock In \emph{International Conference on Machine Learning}, pages
  8748--8763. PMLR.

\bibitem[{Radford et~al.(2018)Radford, Narasimhan, Salimans, Sutskever
  et~al.}]{radford2018improving}
Alec Radford, Karthik Narasimhan, Tim Salimans, Ilya Sutskever, et~al. 2018.
\newblock Improving language understanding by generative pre-training.

\bibitem[{Ramesh et~al.(2021)Ramesh, Pavlov, Goh, Gray, Voss, Radford, Chen,
  and Sutskever}]{Ramesh2021-zk}
Aditya Ramesh, Mikhail Pavlov, Gabriel Goh, Scott Gray, Chelsea Voss, Alec
  Radford, Mark Chen, and Ilya Sutskever. 2021.
\newblock \href {http://arxiv.org/abs/2102.12092} {Zero-shot text-to-image
  generation}.

\bibitem[{Redmon et~al.(2016)Redmon, Divvala, Girshick, and
  Farhadi}]{redmon2016you}
Joseph Redmon, Santosh Divvala, Ross Girshick, and Ali Farhadi. 2016.
\newblock You only look once: Unified, real-time object detection.
\newblock In \emph{Proceedings of the IEEE conference on computer vision and
  pattern recognition}, pages 779--788.

\bibitem[{Russakovsky et~al.(2015)Russakovsky, Deng, Su, Krause, Satheesh, Ma,
  Huang, Karpathy, Khosla, Bernstein et~al.}]{russakovsky2015imagenet}
Olga Russakovsky, Jia Deng, Hao Su, Jonathan Krause, Sanjeev Satheesh, Sean Ma,
  Zhiheng Huang, Andrej Karpathy, Aditya Khosla, Michael Bernstein, et~al.
  2015.
\newblock Imagenet large scale visual recognition challenge.
\newblock \emph{International journal of computer vision}, 115:211--252.

\bibitem[{Sanh et~al.(2019)Sanh, Debut, Chaumond, and
  Wolf}]{sanh2019distilbert}
Victor Sanh, Lysandre Debut, Julien Chaumond, and Thomas Wolf. 2019.
\newblock Distilbert, a distilled version of bert: smaller, faster, cheaper and
  lighter.
\newblock \emph{arXiv preprint arXiv:1910.01108}.

\bibitem[{Schuhmann et~al.(2021)Schuhmann, Vencu, Beaumont, Kaczmarczyk,
  Mullis, Katta, Coombes, Jitsev, and Komatsuzaki}]{schuhmann2021laion}
Christoph Schuhmann, Richard Vencu, Romain Beaumont, Robert Kaczmarczyk,
  Clayton Mullis, Aarush Katta, Theo Coombes, Jenia Jitsev, and Aran
  Komatsuzaki. 2021.
\newblock Laion-400m: Open dataset of clip-filtered 400 million image-text
  pairs.
\newblock \emph{arXiv preprint arXiv:2111.02114}.

\bibitem[{Sennrich et~al.(2015)Sennrich, Haddow, and
  Birch}]{sennrich2015neural}
Rico Sennrich, Barry Haddow, and Alexandra Birch. 2015.
\newblock Neural machine translation of rare words with subword units.
\newblock \emph{arXiv preprint arXiv:1508.07909}.

\bibitem[{Sharma et~al.(2018)Sharma, Ding, Goodman, and
  Soricut}]{sharma2018conceptual}
Piyush Sharma, Nan Ding, Sebastian Goodman, and Radu Soricut. 2018.
\newblock Conceptual captions: A cleaned, hypernymed, image alt-text dataset
  for automatic image captioning.
\newblock In \emph{Proceedings of the 56th Annual Meeting of the Association
  for Computational Linguistics (Volume 1: Long Papers)}, pages 2556--2565.

\bibitem[{Singh et~al.(2022)Singh, Hu, Goswami, Couairon, Galuba, Rohrbach, and
  Kiela}]{singh2022flava}
Amanpreet Singh, Ronghang Hu, Vedanuj Goswami, Guillaume Couairon, Wojciech
  Galuba, Marcus Rohrbach, and Douwe Kiela. 2022.
\newblock Flava: A foundational language and vision alignment model.
\newblock In \emph{Proceedings of the IEEE/CVF Conference on Computer Vision
  and Pattern Recognition}, pages 15638--15650.

\bibitem[{Su et~al.(2019)Su, Zhu, Cao, Li, Lu, Wei, and Dai}]{Su2019-pf}
Weijie Su, Xizhou Zhu, Yue Cao, Bin Li, Lewei Lu, Furu Wei, and Jifeng Dai.
  2019.
\newblock \href {http://arxiv.org/abs/1908.08530} {{VL-BERT}: Pre-training of
  generic {Visual-Linguistic} representations}.

\bibitem[{Suhr et~al.(2018)Suhr, Zhou, Zhang, Zhang, Bai, and
  Artzi}]{suhr2018corpus}
Alane Suhr, Stephanie Zhou, Ally Zhang, Iris Zhang, Huajun Bai, and Yoav Artzi.
  2018.
\newblock A corpus for reasoning about natural language grounded in
  photographs.
\newblock \emph{arXiv preprint arXiv:1811.00491}.

\bibitem[{Tan and Bansal(2019)}]{Tan2019-yi}
Hao Tan and Mohit Bansal. 2019.
\newblock \href {http://arxiv.org/abs/1908.07490} {{LXMERT}: Learning
  {Cross-Modality} encoder representations from transformers}.

\bibitem[{Thomee et~al.(2016)Thomee, Shamma, Friedland, Elizalde, Ni, Poland,
  Borth, and Li}]{thomee2016yfcc100m}
Bart Thomee, David~A Shamma, Gerald Friedland, Benjamin Elizalde, Karl Ni,
  Douglas Poland, Damian Borth, and Li-Jia Li. 2016.
\newblock Yfcc100m: The new data in multimedia research.
\newblock \emph{Communications of the ACM}, 59(2):64--73.

\bibitem[{Touvron et~al.(2021)Touvron, Cord, Douze, Massa, Sablayrolles, and
  J{\'e}gou}]{touvron2021training}
Hugo Touvron, Matthieu Cord, Matthijs Douze, Francisco Massa, Alexandre
  Sablayrolles, and Herv{\'e} J{\'e}gou. 2021.
\newblock Training data-efficient image transformers \& distillation through
  attention.
\newblock In \emph{International Conference on Machine Learning}, pages
  10347--10357. PMLR.

\bibitem[{Vaswani et~al.(2017)Vaswani, Shazeer, Parmar, Uszkoreit, Jones,
  Gomez, Kaiser, and Polosukhin}]{vaswani2017attention}
Ashish Vaswani, Noam Shazeer, Niki Parmar, Jakob Uszkoreit, Llion Jones,
  Aidan~N Gomez, {\L}ukasz Kaiser, and Illia Polosukhin. 2017.
\newblock Attention is all you need.
\newblock \emph{Advances in neural information processing systems}, 30.

\bibitem[{Wang et~al.(2018)Wang, Singh, Michael, Hill, Levy, and
  Bowman}]{Wang2018-fr}
Alex Wang, Amanpreet Singh, Julian Michael, Felix Hill, Omer Levy, and Samuel~R
  Bowman. 2018.
\newblock \href {http://arxiv.org/abs/1804.07461} {{GLUE}: A {Multi-Task}
  benchmark and analysis platform for natural language understanding}.

\bibitem[{Wang et~al.(2022{\natexlab{a}})Wang, Yang, Hu, Li, Lin, Gan, Liu,
  Liu, and Wang}]{wang2022git}
Jianfeng Wang, Zhengyuan Yang, Xiaowei Hu, Linjie Li, Kevin Lin, Zhe Gan,
  Zicheng Liu, Ce~Liu, and Lijuan Wang. 2022{\natexlab{a}}.
\newblock Git: A generative image-to-text transformer for vision and language.
\newblock \emph{arXiv preprint arXiv:2205.14100}.

\bibitem[{Wang et~al.(2023{\natexlab{a}})Wang, Zhou, Shou, and
  Yan}]{wang2023position}
Jinpeng Wang, Pan Zhou, Mike~Zheng Shou, and Shuicheng Yan. 2023{\natexlab{a}}.
\newblock Position-guided text prompt for vision-language pre-training.
\newblock In \emph{Proceedings of the IEEE/CVF Conference on Computer Vision
  and Pattern Recognition}, pages 23242--23251.

\bibitem[{Wang et~al.(2022{\natexlab{b}})Wang, Chen, Wu, Luo, Zhou, Zhao, Xie,
  Liu, Jiang, and Yuan}]{wang2022omnivl}
Junke Wang, Dongdong Chen, Zuxuan Wu, Chong Luo, Luowei Zhou, Yucheng Zhao,
  Yujia Xie, Ce~Liu, Yu-Gang Jiang, and Lu~Yuan. 2022{\natexlab{b}}.
\newblock Omnivl: One foundation model for image-language and video-language
  tasks.
\newblock \emph{arXiv preprint arXiv:2209.07526}.

\bibitem[{Wang et~al.(2023{\natexlab{b}})Wang, Wang, Lin, Bai, Zhou, Zhou,
  Wang, and Zhou}]{wang2023one}
Peng Wang, Shijie Wang, Junyang Lin, Shuai Bai, Xiaohuan Zhou, Jingren Zhou,
  Xinggang Wang, and Chang Zhou. 2023{\natexlab{b}}.
\newblock One-peace: Exploring one general representation model toward
  unlimited modalities.
\newblock \emph{arXiv preprint arXiv:2305.11172}.

\bibitem[{Wang et~al.(2022{\natexlab{c}})Wang, Yang, Men, Lin, Bai, Li, Ma,
  Zhou, Zhou, and Yang}]{Wang2022-gx}
Peng Wang, An~Yang, Rui Men, Junyang Lin, Shuai Bai, Zhikang Li, Jianxin Ma,
  Chang Zhou, Jingren Zhou, and Hongxia Yang. 2022{\natexlab{c}}.
\newblock \href {http://arxiv.org/abs/2202.03052} {{OFA}: Unifying
  architectures, tasks, and modalities through a simple {Sequence-to-Sequence}
  learning framework}.

\bibitem[{Wang et~al.(2023{\natexlab{c}})Wang, Bao, Dong, Bjorck, Peng, Liu,
  Aggarwal, Mohammed, Singhal, Som et~al.}]{wang2023image}
Wenhui Wang, Hangbo Bao, Li~Dong, Johan Bjorck, Zhiliang Peng, Qiang Liu, Kriti
  Aggarwal, Owais~Khan Mohammed, Saksham Singhal, Subhojit Som, et~al.
  2023{\natexlab{c}}.
\newblock Image as a foreign language: Beit pretraining for vision and
  vision-language tasks.
\newblock In \emph{Proceedings of the IEEE/CVF Conference on Computer Vision
  and Pattern Recognition}, pages 19175--19186.

\bibitem[{Wang et~al.(2021)Wang, Yu, Yu, Dai, Tsvetkov, and Cao}]{Wang2021-se}
Zirui Wang, Jiahui Yu, Adams~Wei Yu, Zihang Dai, Yulia Tsvetkov, and Yuan Cao.
  2021.
\newblock \href {http://arxiv.org/abs/2108.10904} {{SimVLM}: Simple visual
  language model pretraining with weak supervision}.

\bibitem[{Wu et~al.(2016)Wu, Schuster, Chen, Le, Norouzi, Macherey, Krikun,
  Cao, Gao, Macherey et~al.}]{wu2016google}
Yonghui Wu, Mike Schuster, Zhifeng Chen, Quoc~V Le, Mohammad Norouzi, Wolfgang
  Macherey, Maxim Krikun, Yuan Cao, Qin Gao, Klaus Macherey, et~al. 2016.
\newblock Google's neural machine translation system: Bridging the gap between
  human and machine translation.
\newblock \emph{arXiv preprint arXiv:1609.08144}.

\bibitem[{Xu et~al.(2021)Xu, Yan, Li, Bi, Huang, Xiao, and Huang}]{xu2021e2e}
Haiyang Xu, Ming Yan, Chenliang Li, Bin Bi, Songfang Huang, Wenming Xiao, and
  Fei Huang. 2021.
\newblock E2e-vlp: end-to-end vision-language pre-training enhanced by visual
  learning.
\newblock \emph{arXiv preprint arXiv:2106.01804}.

\bibitem[{Xu et~al.(2022)Xu, Wu, Rosenman, Lal, Che, and Duan}]{Xu2022-ce}
Xiao Xu, Chenfei Wu, Shachar Rosenman, Vasudev Lal, Wanxiang Che, and Nan Duan.
  2022.
\newblock \href {http://arxiv.org/abs/2206.08657} {{BridgeTower}: Building
  bridges between encoders in {Vision-Language} representation learning}.

\bibitem[{Yang et~al.(2022{\natexlab{a}})Yang, Li, Zhang, Xiao, Liu, Yuan, and
  Gao}]{Yang2022-gd}
Jianwei Yang, Chunyuan Li, Pengchuan Zhang, Bin Xiao, Ce~Liu, Lu~Yuan, and
  Jianfeng Gao. 2022{\natexlab{a}}.
\newblock Unified contrastive learning in image-text-label space.
\newblock In \emph{2022 {IEEE/CVF} Conference on Computer Vision and Pattern
  Recognition ({CVPR})}. IEEE.

\bibitem[{Yang et~al.(2022{\natexlab{b}})Yang, Gan, Wang, Hu, Ahmed, Liu, Lu,
  and Wang}]{Yang2022-hf}
Zhengyuan Yang, Zhe Gan, Jianfeng Wang, Xiaowei Hu, Faisal Ahmed, Zicheng Liu,
  Yumao Lu, and Lijuan Wang. 2022{\natexlab{b}}.
\newblock {UniTAB}: Unifying text and box outputs for grounded
  {Vision-Language} modeling.
\newblock In \emph{Computer Vision -- {ECCV} 2022}, pages 521--539. Springer
  Nature Switzerland.

\bibitem[{Yu et~al.(2022)Yu, Wang, Vasudevan, Yeung, Seyedhosseini, and
  Wu}]{Yu2022-zn}
Jiahui Yu, Zirui Wang, Vijay Vasudevan, Legg Yeung, Mojtaba Seyedhosseini, and
  Yonghui Wu. 2022.
\newblock \href {http://arxiv.org/abs/2205.01917} {{CoCa}: Contrastive
  captioners are {Image-Text} foundation models}.

\bibitem[{Yu et~al.(2018)Yu, Lin, Shen, Yang, Lu, Bansal, and
  Berg}]{DBLP:journals/corr/abs-1801-08186}
Licheng Yu, Zhe Lin, Xiaohui Shen, Jimei Yang, Xin Lu, Mohit Bansal, and
  Tamara~L. Berg. 2018.
\newblock \href {http://arxiv.org/abs/1801.08186} {Mattnet: Modular attention
  network for referring expression comprehension}.
\newblock \emph{CoRR}, abs/1801.08186.

\bibitem[{Yuan et~al.(2021)Yuan, Chen, Chen, Codella, Dai, Gao, Hu, Huang, Li,
  Li, Liu, Liu, Liu, Lu, Shi, Wang, Wang, Xiao, Xiao, Yang, Zeng, Zhou, and
  Zhang}]{Yuan2021-yg}
Lu~Yuan, Dongdong Chen, Yi-Ling Chen, Noel Codella, Xiyang Dai, Jianfeng Gao,
  Houdong Hu, Xuedong Huang, Boxin Li, Chunyuan Li, Ce~Liu, Mengchen Liu,
  Zicheng Liu, Yumao Lu, Yu~Shi, Lijuan Wang, Jianfeng Wang, Bin Xiao, Zhen
  Xiao, Jianwei Yang, Michael Zeng, Luowei Zhou, and Pengchuan Zhang. 2021.
\newblock \href {http://arxiv.org/abs/2111.11432} {Florence: A new foundation
  model for computer vision}.

\bibitem[{Zellers et~al.(2018)Zellers, Bisk, Farhadi, and
  Choi}]{Zellers2018-nm}
Rowan Zellers, Yonatan Bisk, Ali Farhadi, and Yejin Choi. 2018.
\newblock \href {http://arxiv.org/abs/1811.10830} {From recognition to
  cognition: Visual commonsense reasoning}.

\bibitem[{Zeng et~al.(2022)Zeng, Zhang, Li, Wang, Zhang, and
  Zhou}]{Zeng2022-bp}
Yan Zeng, Xinsong Zhang, Hang Li, Jiawei Wang, Jipeng Zhang, and Wangchunshu
  Zhou. 2022.
\newblock \href {http://arxiv.org/abs/2211.12402} {{X$^2$-VLM}: {All-In-One}
  pre-trained model for {Vision-Language} tasks}.

\bibitem[{Zhai et~al.(2022)Zhai, Wang, Mustafa, Steiner, Keysers, Kolesnikov,
  and Beyer}]{zhai2022lit}
Xiaohua Zhai, Xiao Wang, Basil Mustafa, Andreas Steiner, Daniel Keysers,
  Alexander Kolesnikov, and Lucas Beyer. 2022.
\newblock Lit: Zero-shot transfer with locked-image text tuning.
\newblock In \emph{Proceedings of the IEEE/CVF Conference on Computer Vision
  and Pattern Recognition}, pages 18123--18133.

\bibitem[{Zhang et~al.(2021)Zhang, Li, Hu, Yang, Zhang, Wang, Choi, and
  Gao}]{zhang2021vinvl}
Pengchuan Zhang, Xiujun Li, Xiaowei Hu, Jianwei Yang, Lei Zhang, Lijuan Wang,
  Yejin Choi, and Jianfeng Gao. 2021.
\newblock Vinvl: Revisiting visual representations in vision-language models.
\newblock In \emph{Proceedings of the IEEE/CVF Conference on Computer Vision
  and Pattern Recognition}, pages 5579--5588.

\bibitem[{Zhu et~al.(2015)Zhu, Kiros, Zemel, Salakhutdinov, Urtasun, Torralba,
  and Fidler}]{zhu2015aligning}
Yukun Zhu, Ryan Kiros, Rich Zemel, Ruslan Salakhutdinov, Raquel Urtasun,
  Antonio Torralba, and Sanja Fidler. 2015.
\newblock Aligning books and movies: Towards story-like visual explanations by
  watching movies and reading books.
\newblock In \emph{Proceedings of the IEEE international conference on computer
  vision}, pages 19--27.

\end{thebibliography}
\bibliographystyle{acl_natbib}

\appendix



\end{document}